\documentclass{tADR2e}

\usepackage{enumerate}
\usepackage{algorithm,algorithmic}
\usepackage{comment}
\usepackage{CJKutf8}
\usepackage{multirow}
\usepackage{algorithm}

\makeatletter
  \newcommand{\figcaption}[1]{\def\@captype{figure}\caption{#1}}
  \newcommand{\tblcaption}[1]{\def\@captype{table}\caption{#1}}
\makeatother

\begin{document}

\jvol{00} \jnum{00} \jyear{2021} \jmonth{}

%\articletype{GUIDE}

\title{Multiagent Multimodal Categorization for Symbol Emergence: \\Emergent Communication via Interpersonal Cross-modal Inference}

\author{Yoshinobu Hagiwara$^{a}$$^{\ast}$\thanks{$^\ast$Corresponding author. Email: yhagiwara@em.ci.ritsumei.ac.jp
\vspace{6pt}}, Kazuma Furukawa$^{a}$, Akira Taniguchi$^{a}$, and Tadahiro Taniguchi$^{a}$\\
\vspace{6pt}$^{a}${\em{Ritsumeikan University \\ 1-1-1 Noji Higashi, Kusatsu, Shiga 525-8577, Japan}}\\\vspace{6pt}\received{v1.0 released December 2020} }

\maketitle

\begin{abstract}
This paper describes a computational model of multiagent multimodal categorization that realizes emergent communication. We clarify whether the computational model can reproduce the following functions in a symbol emergence system, comprising two agents with different sensory modalities playing a naming game. 
(1) Function for forming a shared lexical system that comprises perceptual categories and corresponding signs, formed by agents through individual learning and semiotic communication between agents. 
(2) Function to improve the categorization accuracy in an agent via semiotic communication with another agent, even when some sensory modalities of each agent are missing.
(3) Function that an agent infers unobserved sensory information based on a sign sampled from another agent in the same manner as cross-modal inference. 
We propose an interpersonal multimodal Dirichlet mixture (Inter-MDM), which is derived by dividing an integrative probabilistic generative model, which is obtained by integrating two Dirichlet mixtures (DMs). The Markov chain Monte Carlo algorithm realizes emergent communication. 
The experimental results demonstrated that Inter-MDM enables agents to form multimodal categories and appropriately share signs between agents. It is shown that emergent communication improves categorization accuracy, even when some sensory modalities are missing. Inter-MDM enables an agent to predict unobserved information based on a shared sign.

\medskip

\medskip

\end{abstract}

\section{Introduction}\label{sec:1}
\begin{CJK}{UTF8}{min}
Humans have the ability to create and share words (e.g., ``apple'') associated with perceptual categories that are formed based on multimodal sensory information~\citep{Barsalou99,SER,SEC}. In semiotics, a word is a type of sign.
The word ``apple" itself is nothing more than a sequence of sounds that are unrelated to the physical and perceptual features of apples. The word ``apple" represents an apple simply because the red fruit has been habitually called ``apple.'' This is called the arbitrariness of symbols in semiotics~\citep{Chandler07}. Humans can invent new words; if we start to use an arbitrary word and share it within a group, the word becomes a meaningful symbol. 
In other words, humans create and share the meaning of the word "apple," which initially has no fixed meaning, via interpretations through perceptual categories that are formed based on multimodal sensory information.
Perceptual category formation depends on the lexical knowledge obtained through learning a language, i.e., a shared symbol system. 
As a result, a human can form and share symbols with another human in a bottom-up manner and understand the meaning of a sign spoken by the other by inferring (or imagining) unseen sensory information.
For example, the color and shape of an object can be inferred by hearing the word ``apple" spoken by others.
Interestingly, even a human with impaired sensory organs can form perceptual categories and share and use signs for communication.
The goal of this study is to develop a computational model that reproduces the phenomena described above.

\begin{figure}[bt]
\centering
\includegraphics[scale=0.8]{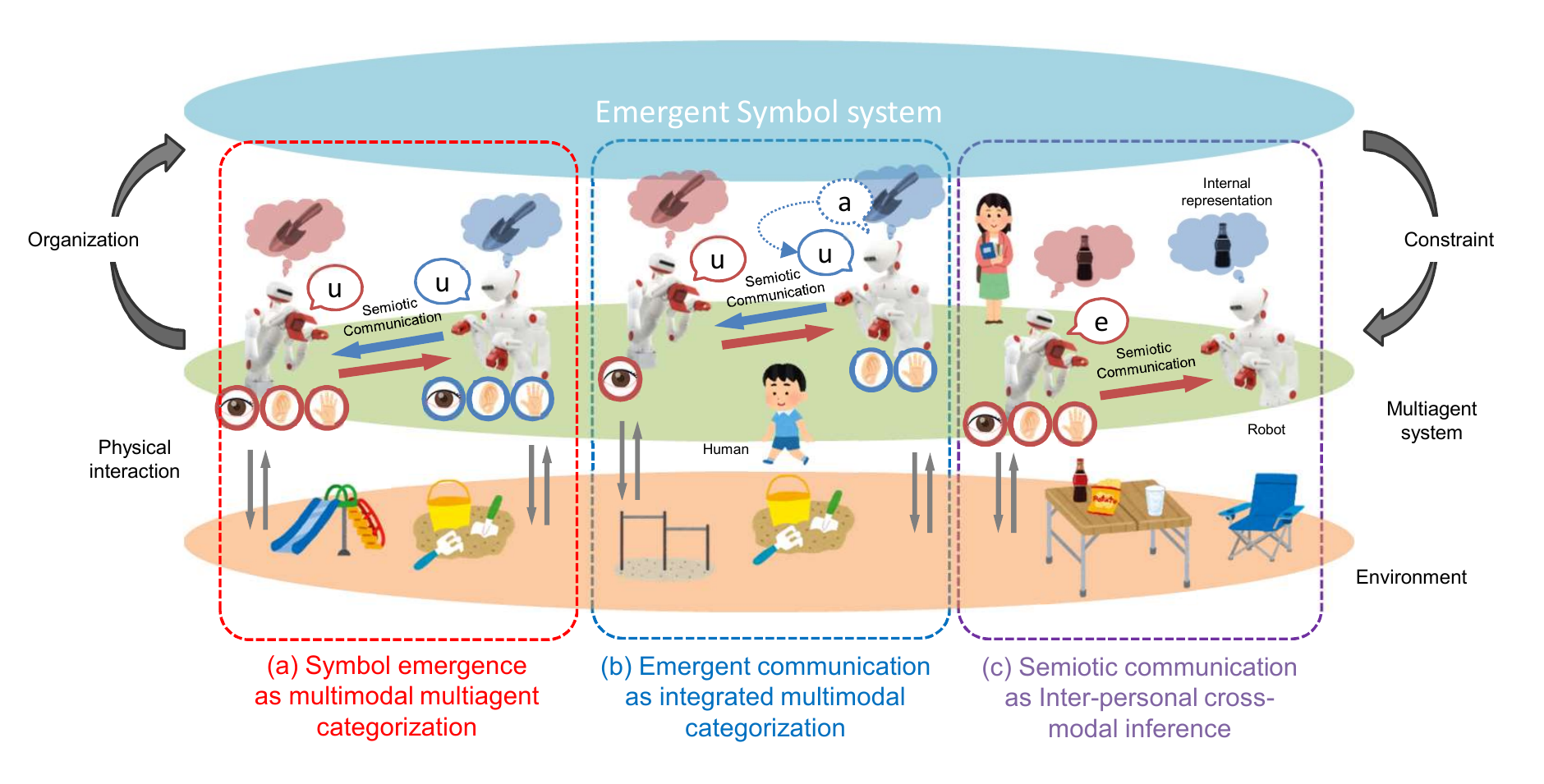}
\caption{Overview of a symbol emergence system where each agent has a multimodal sensory system. (a) Illustration of symbol emergence in a multiagent system in which each agent has a multimodal sensory system. (b) Illustration of the improvement of categorization in an agent by semiotic communication with another agent, even when some modality information is missing. (c) Illustration of the inference of unobserved modality information based on a sign (e.g., u, a, or e) shared between two agents.}
\label{fig:overview}
\end{figure}

To achieve this goal, we constructed a computational model based on a symbol emergence system~\citep{SER,SEC}. 
Fig.~\ref{fig:overview} presents an overview of a symbol emergence system, where each agent has a multimodal sensory system. The symbol emergence system is a multiagent system that can organize an emergent symbol system and enable semiotic communication for each agent. 
We define semiotic communication in the symbol emergence system as the process of interaction between agents, who interpret signs uttered by other agents based on an internal representation (i.e., perceptual categories) and associate these signs with objects.
The symbol emergence system is a complex system with emergent properties. For further details regarding symbol emergence systems, please refer to ~\cite{SER,SEC}. 

Hagiwara et al. proposed a computational model of a symbol emergence system comprising two agents that perform categorization based on a visual modality, i.e., a single modality~\cite{Hagiwara19}. We call the model proposed in~\cite{Hagiwara19} interpersonal Dirichlet mixture (Inter-DM) in this study because the model is obtained by combining two Dirichlet mixtures (DMs). 
Inter-DM is an advanced version of the Talking Heads experiment, in which various computational models of language emergence using perceptual categories based on sensory experiences were proposed by Steels et al.~\citep{Steels15}. In the Talking Heads experiment, the symbol emergence and perceptual categories were decoupled in the models, whereas they were modeled as one learning mechanism in Inter-DM. Specifically, in Inter-DM, emergent communication is modeled as an interpersonal inference based on the Metropolis--Hastings (M-H) algorithm, which is a type of Markov chain Monte Carlo method; this results in symbol emergence by maximizing the marginal likelihood of all agents' observations. In Inter-DM, symbol emergence is theoretically guaranteed by decomposing this marginal likelihood maximization problem into an autonomous decentralized optimization problem with two agents.
However, because Inter-DM was limited to a single modality, the computational model, which achieves symbol emergence based on the categorization of multimodal sensory information, as shown in Fig.~\ref{fig:overview}, has not been clarified.

In a study on object categorization based on multimodal sensory information, Nakamura et al. proposed multimodal latent Dirichlet allocation (MLDA), a probabilistic generative model (PGM) that enables a robot to categorize daily objects (e.g., bottles, cups, and cans) based on multimodal sensory information (e.g., visual, haptic, sound, and language)~\cite{Nakamura09}. This study achieved a bottom-up categorization based on multimodal sensory information. In addition, MLDA enables cross-modal inference in which a robot infers unobserved modality information from the observations of other modalities via a formed category.
However, MLDA did not deal with symbol emergence between agents and cross-modal inference between agents based on a shared sign. Our study describes a model that realizes the emergence of symbols based on multimodal categorization methods.

Regarding symbol emergence systems based on multimodal sensory information, the following three questions arise that have not been verified in previous works~\cite{Hagiwara19,Steels15,Nakamura09}:
\begin{enumerate}
 \item Is it possible to extend Inter-DM~\cite{Hagiwara19} to make it multimodal and realize symbol emergence based on the categorization of multimodal sensory information (as illustrated in Fig.~\ref{fig:overview} (a))?
 \item Is it possible to improve the accuracy of categorization in an agent via semiotic communication with another agent, even if some modalities are missing (as illustrated in Fig.~\ref{fig:overview} (b))?
 \item Is it possible to infer unobserved modality information based on a sign shared between two agents in a bottom-up manner (as illustrated in Fig.~\ref{fig:overview} (c))? 
\end{enumerate}

In this study, we propose an interpersonal multimodal Dirichlet mixture (Inter-MDM) as a computational model to verify the above questions. Inter-MDM is a multimodal extension of an Inter-DM and inherits its properties. The multimodal extension of Inter-DM was conducted by modeling multimodal categorization as multimodal Dirichlet mixtures (MDMs). This extension was inspired by the MLDA~\cite{Nakamura09}. We clarify whether the proposed computational model can realize the functions in the aforementioned questions through experiments using a multimodal dataset acquired by a robot in the real world.

The main contributions of this study are as follows:
\begin{enumerate}
 \item We propose Inter-MDM, which models the symbol emergence system where each agent has multimodal sensory systems, forms multimodal object categories, and shares signs with another agent. In addition, we validate the Inter-MDM on synthetic and real-world data.
 \item We show that Inter-MDM can realize the function of sharing signs associated with categories via semiotic communication even between agents having different sets of modalities (e.g., some of the agents' sensory modalities are impaired) and can even improve the categorization performance. 
 \item We show that Inter-MDM can realize the function that an agent infers unobserved modality information based on a sign uttered by another agent.
\end{enumerate}

The remainder of this paper is organized as follows. Section~\ref{sec:2} describes the background of this study. Section~\ref{sec:3} describes the computational model of Inter-MDM and the inference and prediction algorithms, which can be regarded as a probabilistic naming game. Sections~\ref{sec:4} and~\ref{sec:5} present the experiments using synthetic and real-world data, respectively. Section~\ref{sec:6} describes an experiment that tested the function of semiotic communication as an interpersonal cross-modal inference. Finally, Section~\ref{sec:7} concludes the paper.

\section{Backgrounds}\label{sec:2}
This section describes the background of this study, discussing related work on symbol emergence via language games, multimodal categorization in robotics, and symbol emergence as an interpersonal categorization.

\subsection{Symbol emergence via language games}

In a study on the origin of language, Steels et al. adopted a constructive approach and conducted a series of studies developing computational and robotic models of language evolution. Notably, they performed the Talking Heads experiment~\citep{Steels99, Steels05, Steels15}. In the experiment, it was demonstrated that embodied agents can share new vocabularies by playing language games in a real-world environment. A mechanism for creating a shared vocabulary based on a concept was demonstrated in the setting, wherein cameras attached to robots captured objects with simple colors and shapes (e.g., colored triangles, circles, and rectangles) on a magnetic whiteboard.
Steels also demonstrated that the concepts and vocabularies of spaces emerged between agents solely through bottom-up learning and communication in multiple mobile robots (i.e., AIBO)~\citep{AIBO}. Spranger et al. performed a language game experiment to elucidate the emergence of symbol-grounded spatial languages~\citep{Spranger11, Spranger14} and developed a perceptual system for humanoid robots~\citep{Spranger12}.
The Talking Heads experiment was also improved, considering the complexity of semantics and grammar~\citep{Vogt02,Vogt05,Bleys15,Matuszek18}.
These studies regarding the Talking Heads experiment focused on symbol grounding in language games and built the foundation for constructive studies on language evolution.

However, the experiments were limited to simple objects (e.g., red circles and blue rectangles). Daily objects (e.g., bottles, cups, and cans) with complex shapes and properties found in living environments were not used in the experiments. To form object categories of such objects, multimodal sensory information must be taken into consideration (as shown in \cite{mhdp}).
Additionally, they did not provide an objective function of the overall system. 
Furthermore, previous studies based on language games did not model the probabilistic dependency between perceptual multimodal category formation and symbol emergence, i.e., sharing words. 

Our study addresses category formation based on multimodal information for daily objects.
Our model was developed as a multimodal extension of Inter-DM~\cite{Hagiwara19}, which models the inference algorithm as a naming game based on the M-H algorithm, and it inherits the properties of Inter-DM.
Therefore, it is guaranteed that symbol emergence occurs (i.e., words are invented) to maximize the marginal likelihood (i.e., to predict each agent's observations). 
Our model also addresses the mutual dependency between the formation of the perceptual multimodal category and symbol emergence.

\subsection{Multimodal categorization in robotics}
Studies on unsupervised categorization based on multimodal information obtained by robots have been conducted in the field of symbol emergence in robotics.
Nakamura et al. proposed the MLDA as a PGM that executes object categorization based on multimodal information in a robot~\cite{Nakamura09}. MLDA is an extension of latent Dirichlet allocation (LDA)~\cite{lda}---a PGM for categorizing hidden topics from observed documents---to a model for categorizing hidden object categories from multimodal information (i.e., visual, sound, haptic, and language information) as observations. Experimental results demonstrated that a robot can form object categories close to those formed by humans, based on the multimodal information observed from objects. 

By observing linguistic information, i.e., bag-of-words features, the MLDA can form perceptual categories that are affected by signs provided by human participants. Note that MLDA can deal with the dependency between perceptual categories and symbol systems, i.e., relationships between signs and objects, that people interacting with a robot provide.

In addition, MLDA can estimate unseen observations of a sensory modality from another modality’s sensory observation, for example, predict haptic information from visual information. The estimation process is referred to as cross-modal inference. In MLDA, linguistic input, i.e., a set of words, is regarded as a type of sensory modality. Therefore, recalling a visual image from a word is also modeled as a cross-modal inference. 

Following the success of multimodal categorization using MLDA, many extensions have been proposed~\cite{mhdp,Araki12,Nakamura-Ensemble,mMLDA}. 
Ando et al. proposed hierarchical MLDA to enable a robot to form perceptual categories with a hierarchical structure~\cite{hmlda,Ando13}. Miyazawa et al. constructed a PGM that combines MLDA with a cognitive module that learns grammatical knowledge~\cite{Miyazawa19}.
Taniguchi et al. proposed spatial concept formation methods by applying a similar idea to multimodal information obtained by a mobile robot, including positional information~\cite{spcoa,spcoslam,spcoslam2}.
Hagiwara et al. also proposed a hierarchical spatial category formation model that applies hierarchical MLDA to the inference of a hierarchical structure in spatial categories~\cite{Hagiwara18}. 

These studies have achieved bottom-up categorization based on multimodal sensory information, including linguistic observations. However, these models did not address the emergence of semiotic communications.
They assume a fixed relationship between signs and objects and that the relationships are provided by human participants. In contrast, our study describes a model that realizes the emergence of symbols based on multimodal categorization methods.

\subsection{Symbol emergence as an interpersonal categorization}
A constructive model representing a symbol emergence system, in which individual agents form perceptual categories and share representative signs via semiotic communication between two agents, was proposed~\cite{Hagiwara19}. Here, semiotic communication refers to the exchange of signs as the repeated process of sending a sign associated with a category and receiving a sign from the other agent. 

Inter-DM is a PGM that combines two DMs. Each DM corresponds to an agent. Inter-DM assumes that the visual information of an object jointly observed by two agents is generated from a sign shared between the two agents. Categories in each agent are formed by inferring the latent variables of each DM, and the meanings of signs are shared between the agents by inferring the latent variable representing a shared sign. This study posits that the categorization of objects in each agent and the sharing of signs between agents can be explained as the process of inferring a sign as a latent variable in the model via the M-H algorithm~\cite{Hastings70}. Notably, the entire learning process is regarded as a naming game, and no externally defined rewards are required. Experiments have demonstrated that signs associated with object categories are shared between two agents that only have a visual modality. 

However, owing to the limitation of a single modality in Inter-DM, a computational model that achieves symbol emergence based on multimodal information was not clarified (as described in question (1) in the Introduction). It was also not clarified whether emergent communication improves categorization performance in an agent, even if some modalities are missing (as described in question (2) in the Introduction). Inter-DM can theoretically perform cross-modal inferences between agents based on a shared sign; however, it has not been explicitly verified in Inter-DM (as described in question (3) in the Introduction).

In this study, we develop the Inter-MDM that addresses the symbol emergence between agents with multimodal information by extending the Inter-DM. In addition, we demonstrate that emergent communication improves categorization performance in an agent as a model of integrated multimodal categorization, and semiotic communication can be modeled as an interpersonal cross-modal inference.

\end{CJK}

\section{Inter-MDM}\label{sec:3}
\begin{CJK}{UTF8}{min}
This section describes the Inter-MDM, which is a multimodal extension of the Inter-DM.

\subsection{Generative model}

\begin{figure}[bt]
\begin{tabular}{cc}
  \begin{minipage}{.45\textwidth}
	\centering
	\includegraphics[width=\linewidth]{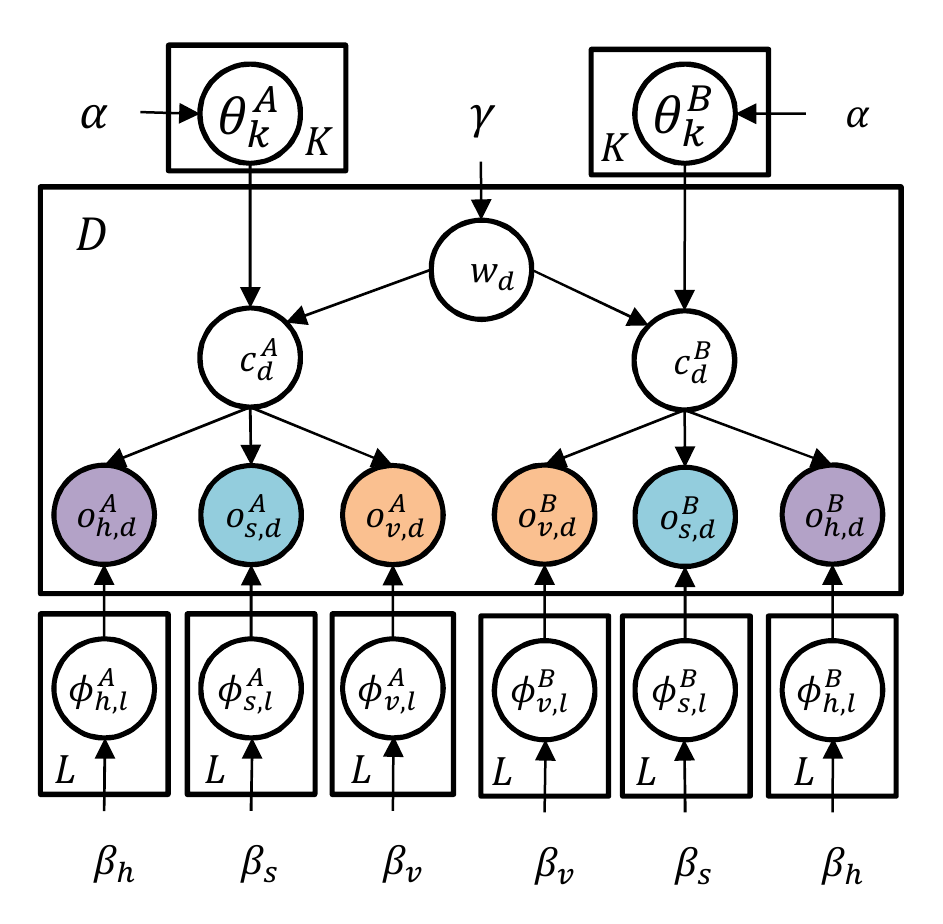}
    \figcaption{Graphical model of the proposed PGM. The generative process of observations in agents A and B is modeled as a PGM, which generates observations from the index of a word ($w_d$) by integrating categories ($c_d^A$, $c_d^B$) of agents A and B.}
    \label{fig:PGM}
  \end{minipage}
  \begin{minipage}{.50\textwidth}
    \begin{center}
    \tblcaption{Definition of variables in the proposed PGM. In the variables $o$ and $\phi$, * represents the modality, i.e., h: haptic, s: sound, and v: visual. $A$ and $B$ depict agents A and B, respectively.}
    \label{tb:PGM}
    \footnotesize
    \renewcommand{\arraystretch}{1.7}
    \begin{tabular}{|c|l|} \hline
      $w_d$ & Index of the sign  \\ \hline
      $c_d^A,c_d^B$ & Index of the category  \\ \hline
      $o_{*,d}^A,o_{*,d}^B$ & Observations\\ \hline
      $\phi_{*,l}^A,\phi_{*,l}^B$ & Parameters of the multinomial distribution \\ \hline
      $\theta_k^A,\theta_k^B$ & Parameters of the categorical distribution\\ \hline
      $\alpha,\beta_*,\gamma$ & Hyperparameters for $\theta,\phi,w_d$ \\ \hline
      $K$ & Number of signs  \\ \hline
      $L$ & Number of categories  \\ \hline
      $D$ & Number of data  points \\ \hline
    \end{tabular}
    \end{center}
  \end{minipage}
\end{tabular}
\end{figure}

A graphical model of the Inter-MDM is illustrated in Fig.~\ref{fig:PGM}. Table~\ref{tb:PGM} presents the definitions of the variables in the proposed PGM. The naming game is obtained as an interpersonal inference procedure based on the M-H algorithm of the generative model in the same manner as in Inter-DM~\cite{Hagiwara19}. 

First, the generative process of the observations of agents A and B is modeled as an integrated PGM. The PGM generates observations ($o_{*,d}^A, o_{*,d}^B$) from the index of a word ($w_d$) by integrating the categories ($c_d^A, c_d^B$) of agents A and B.
The PGM in Fig.~\ref{fig:PGM} can be regarded as a PGM for multimodal categorization based on six modalities by a system that integrates two robots equipped with three modalities. 

The multimodal sensory information ($o_{v,d}$, $o_{s,d}$, and $o_{h,d}$) generated from the index of category ($c_{*,d}$) in each agent is modeled based on the MDM. We assume that the index of a sign $w_{d}$ generates the indices of a category in agents A and B.

The generative process of the proposed model is expressed as:
\begin{eqnarray}
\label{eq:w}
w_d\sim  {\rm Cat}(\gamma)\\
\label{eq:phi_A}
\phi_{*,l}^A\sim {\rm Dir}(\beta_{*})\\
\label{eq:phi_B}
\phi_{*,l}^B\sim {\rm Dir}(\beta_{*})\\
\label{eq:theta_A}
\theta_{k}^A\sim {\rm Dir}(\alpha)\\
\label{eq:theta_B}
\theta_{k}^B\sim {\rm Dir}(\alpha)\\
\label{eq:o_A}
o_{*,d}^A\sim {\rm Multi}(\phi_{*,c_{d}^A}^A)\\
\label{eq:o_B}
o_{*,d}^B\sim {\rm Multi}(\phi_{*,c_{d}^B}^B)\\
\label{eq:c_A}
c_{d}^A\sim {\rm Cat}(\theta_{w_{d}}^A)\\
\label{eq:c_B}
c_{d}^B\sim {\rm Cat}(\theta_{w_{d}}^B),
\end{eqnarray}
where Dir(·), Multi(·), and Cat(·) represent Dirichlet, multinomial, and categorical distributions, respectively. In addition, $\beta$ and $\alpha$ are hyperparameters for Dirichlet distributions, and $d$ is the index of the data. 
In Inter-MDM, multimodal categorization was simply modeled as a multimodal extension of DM~\cite{DM}, which is a generative model that assumes that each data point is generated from a single category. 
When dealing with a model in which one data point is generated from multiple categories, MDM can be extended to MLDA via the multinomial distribution in Equations (\ref{eq:c_A}) and (\ref{eq:c_B}), instead of the categorical distribution. In the graphical model, MDM can be extended to MLDA by adding plates with the number of trials to $c$ and $o$ with the one-hot vector representation.

\subsection{Inference as a naming game}
\label{sec:M-H algorithm}

\begin{figure}[bt]
\centering
\includegraphics[scale=0.75]{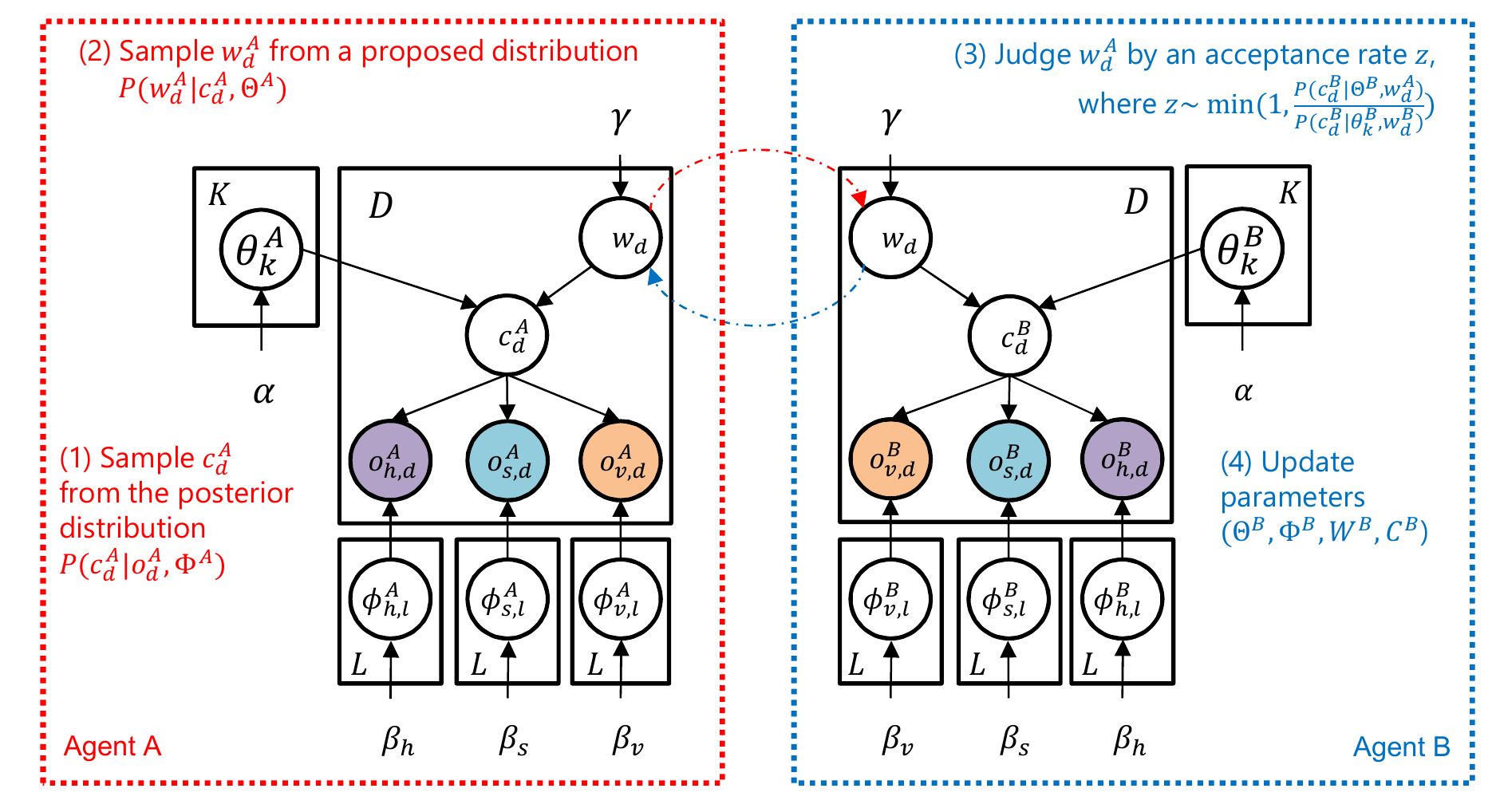}
\caption{Overview of the proposed algorithm for inferring the parameters of the proposed model when agent A is the speaker and agent B is the listener. The proposed algorithm comprises the following steps. (1) Sample $c_d^A$ from the posterior distribution $P(c_d^A|o_{d}^A,{\bf{\Phi}}^A)$. (2) Sample $w_d^A$ from the proposed distribution ($P(w_d^A|c_d^A,\Theta^A)$). (3) Judge $w_d^A$ using acceptance rate $z$. (4) Update the parameters. (5) Switch agents and repeat processes (1) to (4).}
\label{fig:model}
\end{figure}

The inference algorithm of the proposed PGM is described as a naming game~\cite{Steels99} played by two agents, a speaker and a listener, as follows:
\begin{enumerate}
 \item The speaker estimates the category based on the information observed from the object.
 \item The speaker assigns a word based on an estimated category and sends it to the listener.
 \item The listener interprets the word sent by the speaker.
 \item The listener updates the knowledge of words associated with categories.
 \item Switch the speaker and the listener and repeat processes (1) to (4).
\end{enumerate}

The inference algorithm of the proposed PGM can be derived as a naming game based on the M-H algorithm in the same manner as in Inter-DM~\cite{Hagiwara19}. The inference algorithm of Inter-DM was inspired by the Symbol Emergence in Robotics tool KIT (SERKET), a distributed development framework that enables the construction and inference of a large-scale PGM by connecting small-scale PGMs as sub-modules~\cite{SERKET}.

Gibbs sampling and variational inference procedures are well-known inference algorithms for the PGM. The Gibbs sampling algorithm for the PGM is described in Appendix~\ref{sec:Gibbs sampling}.
However, if the Gibbs sampling algorithm is applied to the proposed PGM, the following problem occurs.
In Gibbs sampling, the information of $c^A_d$ and $c^B_d$ should be used simultaneously to sample $w_d$. 
However, the latent variables $c^A_d$ and $c^B_d$ are internal representations of agents A and B, respectively, and a person cannot directly observe another person's internal representation.
Therefore, an inference procedure that requires simultaneous access to the internal representations (i.e., latent variables) of two different agents cannot model a symbol emergence system.

\cite{Hagiwara19} proposed using the M-H algorithm for sampling $w_d$. This approach enables us to divide the PGM into two PGMs corresponding to the two agents. The inference procedure is divided into an interpersonal inference procedure, which can be Gibbs sampling, and interpersonal communication, which is based on the M-H algorithm. The entire procedure can be regarded as a (probabilistic) naming game.
In other words, the naming game can theoretically decompose the inference procedure and derive a decentralized inference algorithm.

As illustrated in Fig.~\ref{fig:model}, the proposed algorithm comprises the following processes. The algorithm is a multimodal extension of that in \cite{Hagiwara19}. 
\begin{enumerate}
 \item Sample the index of a category ($c_d^A$) from the posterior distribution $P(c_d^A|o_{d}^A,{\bf{\Phi}}^A)$ based on the observed multimodal information (i.e., $o_{v,d}^A$, $o_{s,d}^A$, and $o_{h,d}^A$) of agent A.
 \item Sample a sign ($w_d^A$) from the proposed distribution $P(w_d^A|c_d^A,{\bf{\Theta}}^A)$ with the parameters of agent A.
 \item Stochastically accept the sign ($w_d^A$) based on an acceptance ratio ($z$) calculated using the M-H algorithm on the part of agent B.
 \item Update the parameters of agent B if the sign ($w_d^A$) is accepted.
 \item Switch agents and repeat processes (1) to (4).
\end{enumerate}
Steps 1 and 2 can be explained as a process in which agent A suggests a sign ($w_d^A$) based on agent A's observation ($o^A$).
Steps 3 and 4 can be explained as a process in which agent B stochastically accepts the sign ($w_d^A$) based on agent B's categorical knowledge and updates its knowledge based on the suggested sign ($w_d^A$).
These processes, which are performed by switching agents, are interpreted as a naming game based on observations between agents. These processes can be interpreted as emergent communication, in which two agents learn a communication protocol for a categorization task. Emergent communication can be described as the inference of a shared sign, which is an integrated category of two agents performing multimodal categorization.

Algorithm~\ref{alg:alg2} presents the details of the proposed algorithm as a naming game based on the multimodal information between agents A and B. Algorithm~\ref{alg:alg3} presents the internal algorithm of the M-H algorithm function in Algorithm~\ref{alg:alg2}. 
$\bf{W}$, $\bf{C}$, $\bf{\Theta}$, $\bf{\Phi}$, and $\bf{O}$ represent a set of signs, categories in each agent, parameters ($\theta$), parameters ($\phi$), and observations for all data, respectively. $I$ denotes the number of iterations.
$Sp$ and $Li$ represent the speaker and listener, respectively. Unif(·) represents a uniform distribution. In the third line of Algorithm~\ref{alg:alg2}, the sampling and judgment of a set of signs (${\rm \bf{W}}$) are performed with agents A and B as the speaker and listener, respectively. In the fourth line of Algorithm~\ref{alg:alg2},  the sampling and judgment of a set of signs (${\rm \bf{W}}$) are performed with agents B and A as the speaker and listener, respectively. The sampling of a sign ($w_d$) is described in the third line of Algorithm~\ref{alg:alg3}. The sign ($w_d$) is assessed based on the acceptance rate ($z$) described in the fourth line of Algorithm~\ref{alg:alg3}. 

Because the proposed algorithm infers parameters based on the M-H algorithm, sampling from the true posterior distribution of $P(w|o^A, o^B)$ can be performed in the naming game. In other words, symbol emergence is considered a Bayesian inference.

\begin{algorithm}
\caption{Inference as a naming game between two agents}
\label{alg:alg2}
    \begin{algorithmic}[1]
    \small
      \STATE Initialize all parameters
      \FOR{$i = 1$ to $I$}
        \STATE ${\bf W}^{B[i]},{\bf{C}}^{B[i]},{\bf{\Theta}}^{B[i]}$ = 
        \\M-H algorithm(${\bf{C}}^{A[i-1]},{\bf{\Theta}}^{A[i-1]},{\bf{O}}^{B}_{v},{\bf{O}}^{B}_{s},{\bf{O}}^{B}_{h},{\bf W}^{B[i-1]},{\bf{C}}^{B[i-1]},{\bf{\Theta}}^{B[i-1]}$)
        \STATE ${\bf W}^{A[i]},{\bf{C}}^{A[i]},{\bf{\Theta}}^{A[i]}$ = 
        \\M-H algorithm(${\bf{C}}^{B[i]},{\bf{\Theta}}^{B[i]},{\bf{O}}^{A}_{v},{\bf{O}}^{A}_{s},{\bf{O}}^{A}_{h},{\bf W}^{A[i-1]},{\bf{C}}^{A[i-1]},{\bf{\Theta}}^{A[i-1]}$)
      \ENDFOR
    \end{algorithmic}
\end{algorithm}

\begin{algorithm}
\caption{M-H algorithm}
\label{alg:alg3}
\begin{algorithmic}[1]
\small
\STATE M-H algorithm$({\rm \bf{C}}^{Sp},{\rm \bf{\Theta}}^{Sp},{\rm \bf{O}}_v^{Li},{\rm \bf{O}}_s^{Li},{\rm \bf{O}}_h^{Li},{\rm \bf{W}}^{Li},{\rm \bf{C}}^{Li},{\bf{\Theta}}^{Li})$:
\FOR {$d=1$ to $D$}
\STATE $w_{d}^{Sp} \sim P(w_{d}^{Sp}|c_{d}^{Sp},{\bf{\Theta}}^{Sp})$
\STATE $z \sim {\rm min}\left(1,
\dfrac{
P(c_{d}^{Li}|{\bf{\Theta}}^{Li},w_{d}^{Sp})}
{
P(c_{d}^{Li}|{\bf{\Theta}}^{Li},w_{d}^{Li})         
}
\right)$
\STATE $u \sim {\rm Unif}(0,1)$
\IF {$u\leq z$}
\STATE $w_d = w_d^{Sp}$
\ELSE
\STATE $w_d = w_d^{Li}$
\ENDIF
\ENDFOR
\FOR {$l=1$ to $L$}
\STATE $\phi_{v,l}^{Li} \sim {\rm Dir}(\phi_{v,l}^{Li}|{\rm \bf{O}}_v^{Li},{\rm \bf{C}}^{Li},\beta_{v}) $
\STATE $\phi_{s,l}^{Li} \sim {\rm Dir}(\phi_{s,l}^{Li}|{\rm \bf{O}}_s^{Li},{\rm \bf{C}}^{Li},\beta_{s}) $
\STATE $\phi_{h,l}^{Li} \sim {\rm Dir}(\phi_{h,l}^{Li}|{\rm \bf{O}}_h^{Li},{\rm \bf{C}}^{Li},\beta_{h}) $
\ENDFOR
\FOR {$k=1$ to $K$}
\STATE  $\theta_{k}^{Li} \sim {\rm Dir}(\theta_{k}^{Li}|{\rm \bf{C}}^{Li},{\rm \bf{W}},\alpha) $
\ENDFOR
\FOR {$d=1$ to $D$}
\STATE  $c_{d}^{Li} \sim 
{\rm Cat}(c_{d}^{Li}|\theta_k^{Li})
{\rm Multi}(o_{v,d}^{Li}|\phi_{v,l}^{Li})
{\rm Multi}(o_{s,d}^{Li}|\phi_{s,l}^{Li})    
{\rm Multi}(o_{h,d}^{Li}|\phi_{h,l}^{Li}) $
\ENDFOR
\RETURN {${\rm \bf{W}},{\rm \bf{C}}^{Li},{\bf{\Theta}}^{Li}$}
\end{algorithmic} 
\end{algorithm}

\subsection{Semiotic communication as interpersonal cross-modal inference}
Semiotic communication can be modeled as interpersonal cross-modal inference.
In a series of studies on multimodal categorization, inference of unobserved modality sensory information using observed modality sensory information, for example, inferring haptic information from visual information, is called cross-modal inference. Furthermore, in the PGM of Inter-MDM, inferring agent B's observation ($o^A_{*,d}$) from agent A's observation ($o^B_{*,d}$) can be performed in the same manner as cross-modal inference in conventional multimodal PGMs. In this paper, we refer to this inference as {\it interpersonal cross-modal inference}. 

The process models semiotic communication where agent B utters a name, i.e., a sign, of the observed object, and agent A images the sensory information of the sign.  

Interpersonal cross-modal inference enables one agent to predict unobserved information from a sign sampled from another agent based on observations of the other agent and inferred model parameters. Interpersonal cross-modal inference comprises two processes: sampling a sign ($w_{*,d}^A)$ based on the observed information ($o_d^A$) of agent A, and prediction of observation ($\hat{o_{i,d}^B}$) of agent B based on the sampled sign ($w_d^A$) as follows:

\begin{eqnarray}
\label{eq:Cross_1}
c_{d}^{A} &\sim&
P(c_{d}^{A}|o_{v,d}^{A},o_{s,d}^{A},o_{h,d}^{A},{\bf{\Phi}}_{v}^{A},{\bf{\Phi}}_{s}^{A},{\bf{\Phi}}_{h}^{A}) \\
\label{eq:Cross_2}
w_{d}^{A} &\sim& P(w_{d}^{A}|c_{d}^{A},{\bf{\Theta}}^{A}) \\
\label{eq:Cross_3}
c_{d}^{B} &\sim&
P(c_{d}^{B}|w_d^A,{\bf{\Theta}}^{B}) \\
\label{eq:Cross_4}
\hat{o_{v,d}^{B}} &\sim&
P(o_{v,d}^{B}|c_{d}^{B},{\bf{\Phi}}_{v}^{B}) \\
\label{eq:Cross_5}
\hat{o_{s,d}^{B}} &\sim&
P(o_{s,d}^{B}|c_{d}^{B},{\bf{\Phi}}_{s}^{B}) \\
\label{eq:Cross_6}
\hat{o_{h,d}^{B}} &\sim&
P(o_{h,d}^{B}|c_{d}^{B},{\bf{\Phi}}_{h}^{B}).
\end{eqnarray}

\begin{figure}[bt]
\centering
\includegraphics[scale=0.75]{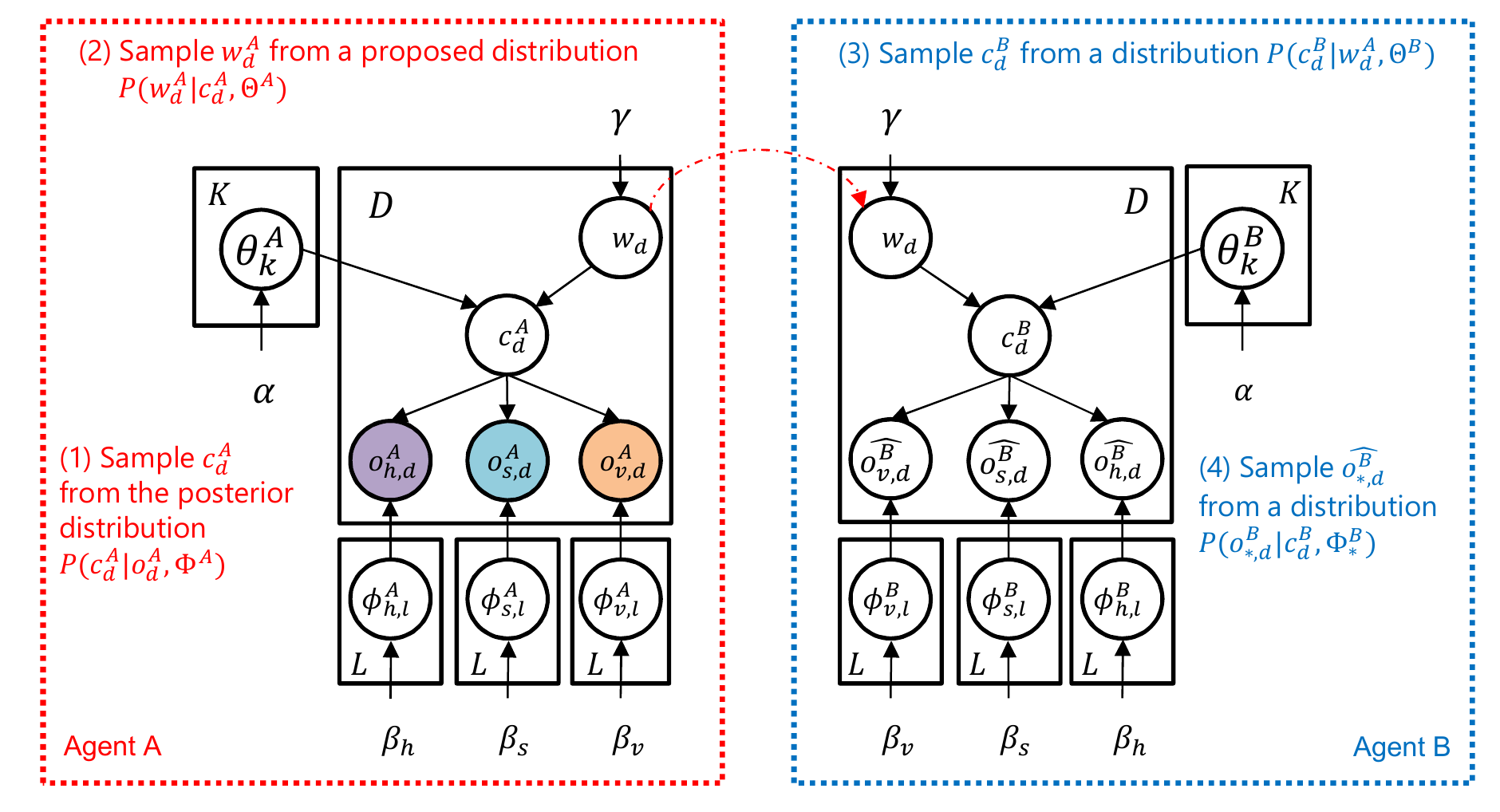}
\caption{Overview of the process of interpersonal cross-modal inference in the proposed PGM. Interpersonal cross-modal inference comprises the following steps. (1) Sample $c_d^A$ from the posterior distribution $P(c_d^A|o_{d}^A,{\bf{\Phi}}^A)$. (2) Sample $w_d^A$ from the proposed distribution $P(w_d^A|c_d^A,\Theta^A)$. (3) Sample $c_d^B$ from the distribution $P(c_d^B|w_d^A,\Theta^B)$. (4) Sample ${\hat o_{*,d}^B}$ from the distribution $P(o_{*,d}^B|c_{d}^B, {\bf{\Phi}}^B)$. Because agent B infers the observation ($o_{*,d}^B$) only from the sign ($w_d^A$) from agent A and its own internal representation, $\Theta^B$ and $\Phi^B$ are fixed, and $o_{*,d}^B$ is unobserved.}
\label{fig:inference}
\end{figure}

In Equation (\ref{eq:Cross_1}), a category ($c_d^A$) is sampled from the posterior probability distribution based on the observed information ($o_{*,d}^A$) and the inferred parameters (${\bf{\Phi}}_*^{A}$) of agent A (as illustrated in Fig.~\ref{fig:inference} (1)). This corresponds to the process in which agent A observes an object and infers the category of the object.
In Equation (\ref{eq:Cross_2}), a sign ($w_d^A$) is sampled by agent A based on the posterior probability distribution, which is based on the sampled category ($c_d^A$) and inferred parameters (${\bf{\Theta}}^{A}$) (as illustrated in Fig.~\ref{fig:inference} (2)). This corresponds to the process of uttering a sign based on inferred categorical knowledge.
In Equation (\ref{eq:Cross_3}), a category ($c_d^B$) of agent B is generated from the probability distribution based on a sign ($\hat{w_d^A}$) sampled from agent A and the parameters (${\bf{\Theta}}^{B}$) inferred in agent B (as illustrated in Fig.~\ref{fig:inference} (3)). This corresponds to the inference of categories based on signs sampled from other agents and their own categorical knowledge.
In Equations (\ref{eq:Cross_4})–-(\ref{eq:Cross_6}), the unobserved information ($\hat{o_{*,d}^{B}}$) of agent B is predicted from the inferred category ($\hat{c_d^B}$) (as illustrated in Fig.~\ref{fig:inference} (4)). This corresponds to the imagination of unobserved information based on the inferred category.
This computational process, an interpersonal cross-modal inference, can be interpreted as a semiotic communication by which an agent infers unobserved modality sensory information based on a sign uttered by another agent.
In addition, this computational process in the PGM is equivalent to the ancestral sampling of $P(o^A|o^B)$.

\end{CJK}

\section{Experiment 1: Synthetic dataset}\label{sec:4}
\label{sec:exp_synthetic}
\begin{CJK}{UTF8}{min}
To verify the hypotheses described in Section~\ref{sec:hypotheses} for Inter-MDM, we performed an experiment using synthetic data.

\subsection{Hypotheses}
\label{sec:hypotheses}
\begin{enumerate}
 \item The model enables the agents to form perceptual categories and share signs based on multimodal sensory information.
 \item The categorization accuracy in an agent is improved via semiotic communication with another agent, even when some modality information is missing.
\end{enumerate}
The hypotheses were verified in different conditions presented in Table~\ref{tab:condition}.
Hypothesis (1) is verified in condition 1, in which two agents have the same three modalities.
Hypothesis (2) is verified in conditions 2--4, with three patterns in each condition.
In condition 2, one agent has three modalities and the other lacks a modality. In condition 3, one agent has three modalities and the other lacks two modalities. In condition 4, one agent lacks one modality and the other lacks two modalities. For example, agent A has visual and sound modalities, whereas agent B has haptic modalities. In this condition, three modalities are completed by integrating the two agents. Under these conditions, we evaluate the similarity of the categories and the sharing of signs between two agents with four communication types. From the experimental results under ideal conditions using a synthetic dataset, we verified hypotheses (1) and (2).
\begin{table}[tb!]
  \begin{center}
  \caption{Conditions and patterns of modalities for each agent in the experiment. Condition 1: Both agents have three modalities. Condition 2: Agent B lacks one modality. Condition 3: Agent B lacks two modalities. Condition 4: There are three modalities by integrating the modalities of agents A and B.}
    \small
    \begin{tabular}{ c c | c c c | c c c} \hline
     & & & Agent A & & & Agent B & \\
    Condition & Pattern & Vision & Sound & Haptic & Vision & Sound & Haptic\\ \hline
    1 : No lack of modalities & - & $\checkmark$ & $\checkmark$ & $\checkmark$ & $\checkmark$ & $\checkmark$ & $\checkmark$\\ \hline
    & I & $\checkmark$ & $\checkmark$ & $\checkmark$ & $\checkmark$ & $\checkmark$ & \\
    2 : Lack of one modality & II & $\checkmark$ & $\checkmark$ & $\checkmark$ &  & $\checkmark$ & $\checkmark$\\
    & III & $\checkmark$ & $\checkmark$ & $\checkmark$ & $\checkmark$ &  & $\checkmark$\\ \hline
    & I & $\checkmark$ & $\checkmark$ & $\checkmark$ & $\checkmark$ &  & \\
    3 : Lack of two modalities & II & $\checkmark$ & $\checkmark$ & $\checkmark$ &  & $\checkmark$ & \\
    & III & $\checkmark$ & $\checkmark$ & $\checkmark$ &  &  & $\checkmark$\\ \hline
    & I & $\checkmark$ & $\checkmark$ &  &  &  & $\checkmark$\\
    4 : Lack of three modalities & II &  & $\checkmark$ & $\checkmark$ & $\checkmark$ &  & \\
    & III & $\checkmark$ &  & $\checkmark$ &  & $\checkmark$ & \\ \hline
    \end{tabular}
  \label{tab:condition}
  \end{center}
\end{table}

\subsection{Conditions}
To verify hypotheses (1) and (2), we evaluated the categorization accuracy and sign sharing in Inter-MDM from comparisons of the following four algorithms including two baselines and a top line.
The proposed algorithm is the inference algorithm of Inter-MDM. All acceptance and all rejection are the baseline algorithms. The integrated model is considered as the top line. The four algorithms can be interpreted as the following types of communication.
\begin{itemize}
    \item Proposed algorithm:\\
    As described in Section~\ref{sec:M-H algorithm}, the proposed algorithm, which accepts sign ($w$) probabilistically based on the acceptance rate ($z$), calculated with formed categories in an agent, is regarded as the communication type of the naming game.
    \item All acceptance (base line 1) :\\
    This is a modified proposed algorithm with $z = 1$, in which the acceptance rate ($z$) of the M-H algorithm in the fifteenth line of Algorithm~\ref{alg:alg3} is set to one. This algorithm is interpreted as a communication that believes and accepts all sign suggestions from other agents and reflects them to form categories. The algorithm is adopted as the communication type of all acceptance as a baseline.
    \item All rejection (base line 2) :\\
    This is the modified proposed algorithm with $z = 0$, in which the acceptance rate ($z$) in the fifteenth line of Algorithm~\ref{alg:alg3} is set to zero. This algorithm is interpreted as a communication that rejects all sign suggestions from other agents and learns categories based solely on its observations. The algorithm is adopted as the communication type of all rejection as a baseline.
    \item Integrated model (top line) :\\
    As described in Section~\ref{sec:Gibbs sampling}, the algorithm that samples from the joint distribution, based on Gibbs sampling, is regarded as a communication type that connects the brains between the agents as a top line.
\end{itemize}
Hypotheses (1) and (2) are verified by comparing the experimental results with the four communication types in conditions 1 to 4.

As evaluation criteria in the experiment, the adjusted Rand index (ARI)~\cite{Hubert85} was adopted to evaluate the results of categorization in each agent, and the Kappa coefficient~\cite{Cohen60} was adopted to evaluate the sign sharing between agents.

The ARI was calculated using the following equation:
\begin{eqnarray}
\label{eq:ari}
ARI &=& \frac{{\rm RI-Expected\:RI}}{{\rm \max (RI)-Expected\:RI}},
\end{eqnarray}
where RI is the Rand Index.

The Kappa coefficient ($\kappa$) was calculated using the following equation:
\begin{eqnarray}
\label{eq:kp}
\kappa &=& \frac{C_o - C_e}{1 - C_e},
\end{eqnarray}
where $C_o$ is the coincidence rate of signs between agents, and $C_e$ is the coincidence rate of signs between agents by random chance. The $\kappa$ value is judged as follows: ($0.81-1.00$) as almost perfect agreement, ($0.61-0.80$) as substantial agreement, ($0.41-0.60$) as moderate agreement, ($0.21-0.40$) as fair agreement, ($0.00-0.20$) as slight agreement, and ($\kappa<0.0$) as no agreement~\citep{Criteria}.

As an experiment in an ideal environment, we verified the questions using a synthetic dataset constructed in a pseudo manner, based on the multimodal symbol emergence model. The synthetic dataset consists of a set of observed data for 15 types of pseudo-objects, which comprise three types of information for the modalities of vision, sound, and haptics. The observed data for each modality are represented by a twenty-dimensional feature histogram.
The hyperparameters of the computational model for the experiment were set as follows: $\alpha=0.01$, $\beta_v=0.001$, $\beta_s=0.001$, and $\beta_h=0.001$. The number of data points $D$ was 150 (10 data points for 15 objects). The number of categories and signs were set as $K=15$ (a,b,c,d,e,f,g,h,i,j,k,l,m,n,o) and $L=15$, respectively.

\subsection{Experimental results}
\subsubsection{Results in condition 1}
Table~\ref{tab:syn_ex_1} presents the experimental results for condition 1, i.e., both agents have three modalities. The ARIs of agents A and B for evaluating the accuracy of object categorization, and the kappa coefficients for evaluating the sharing of signs between two agents, are described with means and standard deviations for four communication types, including the proposed algorithm, all rejection, all acceptance, and the integrated model.
A t-test was performed to compare the proposed algorithm to other communication types; the results of the t-test are described in Table~\ref{tab:syn_ex_1}.
First, in the results of the proposed algorithm and all acceptance, a significant difference $(p < 0.01)$ was confirmed in the ARIs of object categories for each agent. In terms of the kappa coefficient ($\kappa$), which is a criterion for sharing signs between agents, the $\kappa$ value of all rejection was in $(0.61-0.8)$, indicating substantial agreement, whereas the $\kappa$ value of the proposed algorithm was in $(0.81-1.00)$, indicating almost perfect agreement.
Next, in the results of the proposed algorithm and all rejection, no significant difference was determined in the ARIs of object categories for each agent. The $\kappa$ value of all rejection was $(\kappa < 0.0)$, indicating no agreement.
Finally, the ARIs of object categories in the proposed algorithm were not significantly different from those of the integrated model (Gibbs sampling) as the top line. The kappa coefficient of the integrated model cannot be calculated because Gibbs sampling does not sample from the proposed distribution.

\begin{table}[htbp]
\centering
\caption{Results of condition 1 with the ARI means and standard deviations and the means of Kappa coefficients ($\kappa$) in 10 trials with 200 iterations.
In the t-test, **: $(p<0.01)$, *: $(p<0.05)$, n.s.: not significant. The largest and second largest values in the ARIs and Kappa coefficients without the top line are indicated as bold-underlined and bold, respectively.}
\scalebox{0.80}{
\begin{tabular}{c|l|c|c|l|c|c|l|c|c}
\hline
\multicolumn{2}{c|}{Condition 1} & \multicolumn{3}{c|}{ARI (Agent A)} & \multicolumn{3}{c|}{ARI (Agent B)} & \multicolumn{1}{c|}{$\kappa$} & \multicolumn{1}{c}{ARI (W)}\\ \hline \hline
Pattern & Communication types & Mean & SD & t-test & Mean & SD & t-test & Mean & Mean \\ \hline
\multirow{4}{*}{\shortstack{I: \\Agent A (VSH) \\Agent B (VSH)}} & Proposed algorithm& \textbf{0.90} & 0.03 & - & \textbf{0.90} & 0.03 & - & \textbf{\underline{0.97}} & 0.93 \\ 
& All acceptance & 0.82 & 0.05 & ** & 0.83 & 0.05 & ** & \textbf{0.63} & - \\
& All rejection & \textbf{\underline{0.92}} & 0.05 & n.s. & \textbf{\underline{0.91}} & 0.04 & n.s. & -0.00 & - \\ \cline{2-10}
& Integrated model (top line) & 0.89 & 0.03 & n.s. & 0.90 & 0.05 & n.s. & - & - \\ \hline
\end{tabular}
}
\label{tab:syn_ex_1}
\end{table}

\subsubsection{Results in conditions 2--4}
Table~\ref{tab:syn_ex_2} presents the experimental results for condition 2, i.e., one modality of agent B is missing. Under this condition, the proposed algorithm was determined to be significantly different in terms of the ARI of object categorization from other communication types. In particular, a significant difference $(p < 0.01)$ was verified compared to the result of all rejection, which corresponds to the categorization by one agent. The proposed algorithm also exhibited almost perfect agreement with respect to the kappa coefficient.

Table~\ref{tab:syn_ex_3} presents the experimental results for condition 3 in Table~\ref{tab:condition}.
Condition 3 is a case in which two modalities of agent B are missing. Under this condition, the proposed algorithm was determined to be significantly different for all rejection, but not for all acceptance and the integrated model in terms of the ARI of object categorization in agent B. The ARIs of the categories in agent B exhibited the same trend as in condition 1. The proposed algorithm maintains a high kappa coefficient, similar to the result of condition 1.

Table~\ref{tab:syn_ex_4} presents the experimental results for condition 4 in Table~\ref{tab:condition}.
Condition 4 is a case in which two modalities of agent B and one modality of agent A are missing. Under this condition, the proposed algorithm was determined to be significantly different for all acceptance and all rejection. It was not significantly different for the integrated model as the top line in terms of the ARI of object categorization in agent A. 

\begin{table}[htbp]
\centering
\caption{Results of condition 2 with the ARI means and standard deviations and the means of Kappa coefficients ($\kappa$) in 10 trials with 200 iterations.
In the t-test, **: $(p<0.01)$, *: $(p<0.05)$, n.s.: not significant. The largest and second largest values in the ARIs and Kappa coefficients without the top line are indicated as bold-underlined and bold, respectively. In the patterns, V, S, and H indicate visual, sound, and haptic modalities, respectively.}
\scalebox{0.8}{
\begin{tabular}{c|l|c|c|l|c|c|l|c|c}
\hline
\multicolumn{2}{c|}{Condition 2} & \multicolumn{3}{c|}{ARI (Agent A)} & \multicolumn{3}{c|}{ARI (Agent B)} & \multicolumn{1}{c|}{$\kappa$} & \multicolumn{1}{c}{ARI (W)}\\ \hline \hline
Pattern & Communication types & Mean & SD & T-text & Mean & SD & T-text & Mean & Mean \\ \hline
\multirow{4}{*}{\shortstack{I: \\Agent A (VSH) \\Agent B (VS)}} & Proposed algorithm & \textbf{0.90} & 0.03 & - & \textbf{\underline{0.72}} & 0.02 & - & \textbf{\underline{0.96}} & 0.85 \\ 
& All acceptance & 0.83 & 0.06 & ** & \textbf{0.69} & 0.03 & * & \textbf{0.54} & - \\
& All rejection & \textbf{\underline{0.91}} & 0.04 & n.s. & 0.67 & 0.02 & ** & 0.01 & - \\ \cline{2-10}
& Integrated model (top line) & 0.89 & 0.04 & n.s. & 0.71 & 0.02 & n.s. & - & - \\ \hline
\multirow{4}{*}{\shortstack{II: \\Agent A (VSH) \\Agent B (SH)}} & Proposed algorithm & \textbf{0.91} & 0.04 & - & \textbf{\underline{0.72}} & 0.02 & - & \textbf{\underline{0.95}} & 0.88 \\ 
& All acceptance & 0.83 & 0.04 & ** & \textbf{0.69} & 0.04 & * & \textbf{0.51} & - \\
& All rejection & \textbf{\underline{0.91}} & 0.05 & n.s. & 0.67 & 0.03 & ** & -0.00 & - \\ \cline{2-10}
& Integrated model (top line) & 0.89 & 0.03 & n.s. & 0.72 & 0.03 & n.s. & - & - \\ \hline
\multirow{4}{*}{\shortstack{III: \\Agent A (VSH) \\Agent B (VH)}} & Proposed algorithm & \textbf{\underline{0.91}} & 0.04 & - & \textbf{\underline{0.71}} & 0.01 & - & \textbf{\underline{0.95}} & 0.89 \\ 
& All acceptance & 0.84 & 0.04 & ** & \textbf{0.68} & 0.04 & * & \textbf{0.52} & - \\
& All rejection & \textbf{0.90} & 0.03 & n.s. & 0.67 & 0.01 & ** & -0.00 & - \\ \cline{2-10}
& Integrated model (top line) & 0.89 & 0.03 & n.s. & 0.72 & 0.03 & n.s. & - & - \\ \hline
\end{tabular}
}
\label{tab:syn_ex_2}
\end{table}

\begin{table}[htbp]
\centering
\caption{Results of condition 3 with the ARI means and standard deviations and the means of Kappa coefficients ($\kappa$) in 10 trials with 200 iterations.
In the t-test, **: $(p<0.01)$, *: $(p<0.05)$, n.s.: not significant. The largest and second largest values in the ARIs and Kappa coefficients without the top line are indicated as bold-underlined and bold, respectively. In the patterns, V, S, and H indicate visual, sound, and haptic modalities, respectively.}
\scalebox{0.8}{
\begin{tabular}{c|l|c|c|l|c|c|l|c|c}
\hline
\multicolumn{2}{c|}{Condition 3} & \multicolumn{3}{c|}{ARI (Agent A)} & \multicolumn{3}{c|}{ARI (Agent B)} & \multicolumn{1}{c|}{$\kappa$} & \multicolumn{1}{c}{ARI (W)}\\ \hline \hline
Pattern & Communication types & Mean & SD & t-test & Mean & SD & t-test & Mean & Mean \\ \hline
\multirow{4}{*}{\shortstack{I: \\Agent A (VSH) \\Agent B (V)}} & Proposed algorithm & \textbf{0.89} & 0.03 & - & \textbf{\underline{0.46}} & 0.03 & - & \textbf{\underline{0.92}} & 0.72 \\ 
& All acceptance & 0.85 & 0.04 & ** & \textbf{0.45} & 0.04 & n.s. & \textbf{0.36} & - \\
& All rejection & \textbf{\underline{0.91}} & 0.04 & n.s. & 0.35 & 0.02 & ** & 0.00 & - \\ \cline{2-10}
& Integrated model (top line) & 0.89 & 0.03 & n.s. & 0.46 & 0.03 & n.s. & - & - \\ \hline
\multirow{4}{*}{\shortstack{II: \\Agent A (VSH) \\Agent B (S)}} & Proposed algorithm & \textbf{\underline{0.91}} & 0.04 & - & \textbf{\underline{0.45}} & 0.03 & - & \textbf{\underline{0.94}} & 0.77 \\ 
& All acceptance & 0.83 & 0.03 & ** & \textbf{0.45} & 0.04 & n.s. & \textbf{0.30} & - \\
& All rejection & \textbf{0.90} & 0.03 & n.s. & 0.33 & 0.01 & ** & -0.00 & - \\ \cline{2-10}
& Integrated model (top line) & 0.89 & 0.03 & n.s. & 0.45 & 0.04 & n.s. & - & - \\ \hline
\multirow{4}{*}{\shortstack{III: \\Agent A (VSH) \\Agent B (H)}} & Proposed algorithm & \textbf{\underline{0.91}} & 0.03 & - & \textbf{\underline{0.46}} & 0.03 & - & \textbf{\underline{0.93}} & 0.70 \\ 
& All acceptance & 0.83 & 0.05 & ** & \textbf{0.45} & 0.03 & n.s. & \textbf{0.30} & - \\
& All rejection & \textbf{0.90} & 0.03 & n.s. & 0.34 & 0.02 & ** & -0.00 & - \\ \cline{2-10}
& Integrated model (top line) & 0.90 & 0.03 & n.s. & 0.45 & 0.03 & n.s. & - & - \\ \hline
\end{tabular}
}
\label{tab:syn_ex_3}
\end{table}

\begin{table}[htbp]
\centering
\caption{Results of condition 4 with the ARI means and standard deviations and the means of Kappa coefficients ($\kappa$) in 10 trials with 200 iterations.
In the t-test, **: $(p<0.01)$, *: $(p<0.05)$, n.s.: not significant. The largest and second largest values in the ARIs and Kappa coefficients without the top line are indicated as bold-underlined and bold, respectively. In the patterns, V, S, and H indicate visual, sound, and haptic modalities, respectively.}
\scalebox{0.8}{
\begin{tabular}{c|l|c|c|l|c|c|l|c|c}
\hline
\multicolumn{2}{c|}{Condition 4} & \multicolumn{3}{c|}{ARI (Agent A)} & \multicolumn{3}{c|}{ARI (Agent B)} & \multicolumn{1}{c|}{$\kappa$} & \multicolumn{1}{c}{ARI (W)}\\ \hline \hline
Pattern & Communication type & Mean & SD & T-text & Mean & SD & t-test & Mean & Mean \\ \hline
\multirow{4}{*}{\shortstack{I: \\Agent A (VS) \\Agent B (H)}} & Proposed algorithm & \textbf{\underline{0.71}} & 0.01 & - & \textbf{\underline{0.47}} & 0.04 & - & \textbf{\underline{0.94}} & 0.72 \\ 
& All acceptance & \textbf{0.69} & 0.03 & * & \textbf{0.45} & 0.02 & n.s. & \textbf{0.17} & - \\
& All rejection & 0.66 & 0.02 & ** & 0.34 & 0.02 & ** & -0.02 & - \\ \cline{2-10}
& Integrated model (top line) & 0.72 & 0.02 & n.s. & 0.46 & 0.04 & n.s. & - & - \\ \hline
\multirow{4}{*}{\shortstack{II: \\Agent A (SH) \\Agent B (V)}} & Proposed algorithm & \textbf{\underline{0.71}} & 0.01 & - & \textbf{\underline{0.46}} & 0.04 & - & \textbf{\underline{0.94}} & 0.68 \\ 
& All acceptance & \textbf{0.69} & 0.03 & * & \textbf{0.45} & 0.02 & n.s. & \textbf{0.20} & - \\
& All rejection & 0.66 & 0.02 & ** & 0.34 & 0.01 & ** & 0.00 & - \\ \cline{2-10}
& Integrated model (top line) & 0.71 & 0.01 & n.s. & 0.47 & 0.04 & n.s. & - & - \\ \hline
\multirow{4}{*}{\shortstack{III: \\Agent A (VH) \\Agent B (S)}} & Proposed algorithm & \textbf{\underline{0.72}} & 0.02 & - & \textbf{\underline{0.44}} & 0.02 & - & \textbf{\underline{0.95}} & 0.71 \\ 
& All acceptance & \textbf{0.69} & 0.02 & ** & \textbf{0.44} & 0.01 & n.s. & \textbf{0.17} & - \\
& All rejection & 0.66 & 0.02 & ** & 0.33 & 0.02 & ** & -0.01 & - \\ \cline{2-10}
& Integrated model (top line) & 0.71 & 0.02 & n.s. & 0.45 & 0.02 & n.s. & - & - \\ \hline
\end{tabular}
}
\label{tab:syn_ex_4}
\end{table}

\begin{comment}
\begin{figure}[htbp]
\begin{center}			
\includegraphics[scale=0.6]{image/sign_example.eps}
\caption{Sampling results of signs for three example objects. The sampling results are described as the 1st, 2nd, and 3rd signs in ten sampled signs for each object. The rate of a sign for ten sampled signs is described in (·). }
\label{fig:sign_example}
\end{center}
\end{figure}
\end{comment}

\subsection{Discussion}
\subsubsection{Condition 1}
In condition 1, we verified hypothesis (1) for Inter-MDM, that the model enables the agents to form perceptual categories and share signs based on multimodal sensory information.
From the experimental results of the ARIs for agents A and B, shown in Table~\ref{tab:syn_ex_1}, it was clarified that Inter-MDM can form perceptual categories with the same high accuracy as the top line.
It was also clarified from the value of $\kappa$ in Table~\ref{tab:syn_ex_1} that Inter-MDM obtained the highest value compared with the baseline algorithms in the sharing of signs.
These experimental results are consistent with those of Inter-DM, and it can be determined that a multimodal extension that retains the characteristics of Inter-DM as a model of the symbol emergence system was achieved.

\subsubsection{Conditions 2 to 4}
In conditions 2 to 4, we verified hypothesis (2) for Inter-MDM that the accuracy of categorization in an agent is improved by semiotic communication with another agent, even when some modality information is missing.
In Table~\ref{tab:syn_ex_2} for condition 2, where one modality of agent B is missing, it can be confirmed that Inter-MDM obtains a higher ARI (agent B) than all rejection, in which categories are formed without communication. 
In Table~\ref{tab:syn_ex_3} for condition 3, in which two modalities of agent B are missing, the difference in ARI (agent B) between Inter-MDM and all rejection is even larger than that in condition 1.
In Table~\ref{tab:syn_ex_4} for condition 4, in which modalities of agents A and B are missing, it can be confirmed through comparison with all rejection that the ARIs of both agents were improved in Inter-MDM.
From these results, it was clarified that the accuracy of categorization in an agent is improved via semiotic communication with another agent even when some modality information is missing.

In pattern 1 in Table~\ref{tab:syn_ex_4}, the ARI (W) of the sign as a joint category between agents is 0.85. This value is higher than the ARI (agent B) of 0.67, which is the result of categorization solely by agent B in all rejection. In the proposed algorithm, because agent B categorized under the influence of sign W, it is presumed that agent B obtained an ARI of 0.72, which was significantly higher than that of all rejection. This can be interpreted as a phenomenon in which top-down constraints from an emergent symbol system improve the accuracy of bottom-up categorization in each agent.

\end{CJK}

\section{Experiment 2: Real-world dataset}\label{sec:5}
\begin{CJK}{UTF8}{min}

To verify the hypotheses described in Section~\ref{sec:hypotheses} in further detail, we performed an experiment using real-world datasets collected from sensors attached to a robot with real objects.

\subsection{Setting for real-world dataset}
\begin{figure}[ht]
  \begin{center}			
  \includegraphics[scale=0.6]{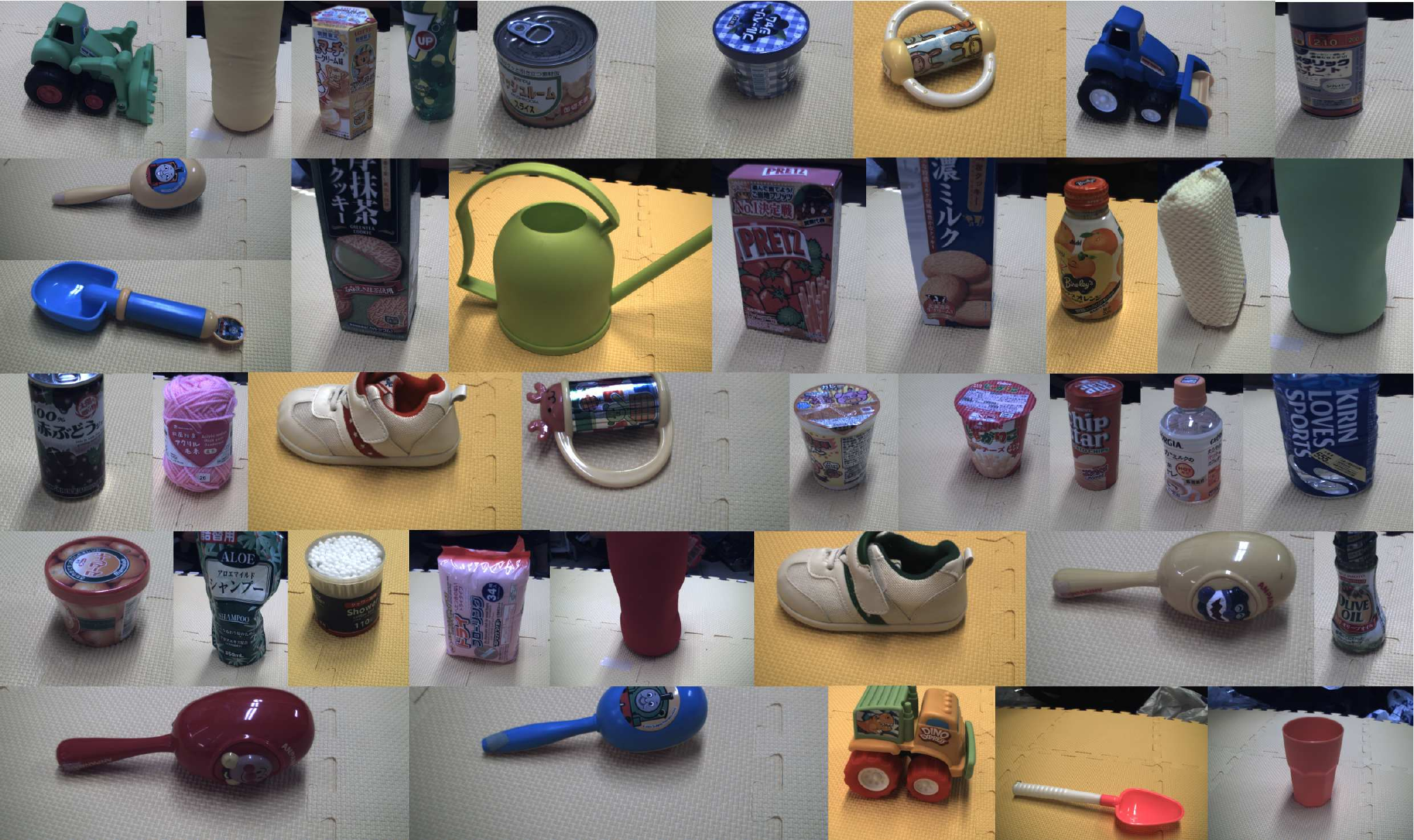}
  \caption{Example images of 40 objects used in the experiment that are selected from the Multimodal Dataset 165~\cite{Dataset}}
  \label{fig:obj}
  \end{center}
\end{figure}

We used the Multimodal Dataset 165~\cite{Dataset}, a set of observed information comprising visual, sound, haptic, and word information, acquired by an arm robot equipped with a vision sensor, pressure sensors, and microphones for 165 types of objects in the real world.
In the experiment, we created two datasets for agents A and B using visual, sound, and haptic information, omitting the word information from this dataset.
\begin{itemize}
 \item
  The visual information in the dataset comprises seven images captured by a CCD camera attached to the robot for each object. In the experiment, two images with different angles were selected from the dataset as images captured by agents A and B. The SIFT features~\cite{sift} comprising 128-dimensional feature vectors were extracted from a captured image, and the feature vectors were converted into a 15-dimensional histogram via k-means clustering and used as observations of visual information.
 \item
 The sound information in the dataset is the sound generated by shaking an object, recorded using a microphone attached to the robot. Random noise was introduced to the sound to create two sound data samples that were adopted as the sound acquired by agents A and B. The 13-dimensional mel-frequency cepstral coefficient (MFCC) was extracted from the acquired sound data as a feature vector, and the feature vector was converted into a 15-dimensional histogram via k-means clustering and used as observations of sound information.
 \item
 The haptic information in the dataset is the time-series data of the sensor values acquired from the 32 pressure sensors when the object is gripped by the robot's hand. Three sets of pressure data, obtained by grasping each object three times, were prepared as 32-dimensional feature vectors. In the experiment, 64-dimensional feature vectors were created by integrating the two 32-dimensional feature vectors for each object as the pressure data acquired by agents A and B. The acquired 64-dimensional features converted into 15-dimensional histograms via k-means clustering were adopted as observations of haptic information.
\end{itemize}
In this experiment, 40 data types were selected from the created datasets (165 types of objects) and used. The objects used in the experiment are shown in Fig.~\ref{fig:obj}.
The hyperparameters for the experiment were set as follows: $\alpha=0.01$, $\beta_v=0.001$, $\beta_s=0.001$, and $\beta_h=0.001$. The number of data points $D$ was 400 (40 types for 10 data points). The numbers of categories and signs were set as $K=40$ and $L=40$, respectively.

\begin{comment}

\begin{figure}[ht]
\begin{center}			
\includegraphics[scale=0.9]{image/robo_1.eps}
\caption{Robot used to create the dataset. [\cite{Dataset}]}
\label{fig:robo_1}
\end{center}
\end{figure}

\begin{figure}[ht]
\begin{center}			
\includegraphics[scale=1.5]{image/robo_2.eps}
\caption{Acquisition of multimodal information. (left) Visual information. (center) Haptic information. (right) Sound information. [\cite{Dataset}]}
\label{fig:robo_2}
\end{center}
\end{figure}

\end{comment}

\subsection{Experimental Results}
Tables~\ref{tab:real_ex_1}, \ref{tab:real_ex_2}, \ref{tab:real_ex_3}, and \ref{tab:real_ex_4} present the experimental results using real-world datasets for conditions 1, 2, 3, and 4, respectively. The conditions of the experiment and the structure of the table are the same as for the experimental results using the synthetic dataset.
In the results of condition 1, we verified the results of the same tendency as in the experiment with the synthetic dataset.
In the results of condition 2, knowledge acceptance was verified to be significantly different from all acceptance in terms of the ARI of categories in agent B. However, no significant difference from all rejection was confirmed.
In the results of condition 3, the proposed algorithm was significantly different from all acceptance and all rejection in terms of the ARI of categories in agent B. 
In the results of condition 4, the proposed algorithm was significantly different from all acceptance and all rejection in terms of the ARI of categories in agent B. In terms of the ARI of categories in agent A, the proposed algorithm was significantly different from all acceptance and all rejection in pattern 3.
The proposed algorithm exhibited almost perfect agreement with respect to the kappa coefficient under all conditions with three patterns.

\begin{table}[bp]
\centering
\caption{Results of condition 1 with the ARI means and standard deviations and the means of Kappa coefficients ($\kappa$) in 10 trials with 300 iterations. In the t-test, **: $(p<0.01)$, *: $(p<0.05)$, n.s.: not significant. The largest and second largest values in the ARIs and Kappa coefficients without the top line are indicated in bold-underlined and bold, respectively.}
\scalebox{0.8}{
\begin{tabular}{c|l|c|c|l|c|c|l|c|c}
\hline
\multicolumn{2}{c|}{Condition 1} & \multicolumn{3}{c|}{ARI (Agent A)} & \multicolumn{3}{c|}{ARI (Agent B)} & \multicolumn{1}{c|}{$\kappa$} & \multicolumn{1}{c}{ARI (W)}\\ \hline \hline
Pattern & Communication type & Mean & SD & t-test & Mean & SD & t-test & Mean & Mean \\ \hline
\multirow{4}{*}{\shortstack{I: \\Agent A (VSH) \\Agent B (VSH)}}& Proposed algorithm & \textbf{0.86} & 0.03 & - & \textbf{0.86} & 0.03 & - & \textbf{\underline{0.98}} & 0.92 \\ 
& All acceptance & 0.23 & 0.11 & ** & 0.25 & 0.10 & ** & \textbf{0.33} & - \\
& All rejection & \textbf{\underline{0.87}} & 0.02 & n.s. & \textbf{\underline{0.86}} & 0.03 & n.s. & -0.00 & - \\ \cline{2-10}
& Integrated model (top line) & 0.87 & 0.02 & n.s. & 0.86 & 0.03 & n.s. & - & - \\ \hline
\end{tabular}
}
\label{tab:real_ex_1}
\end{table}

\begin{table}[bp]
\centering
\caption{Results of condition 2 with the ARI means and standard deviations and the means of Kappa coefficients ($\kappa$) in 10 trials with 300 iterations.
In the t-test, **: $(p<0.01)$, *: $(p<0.05)$, n.s.: not significant. The largest and second largest values in the ARIs and Kappa coefficients without the top line are indicated in bold-underlined and bold, respectively. In the patterns, V, S, and H indicate visual, sound, and haptic modalities, respectively.}
\scalebox{0.8}{
\begin{tabular}{c|l|c|c|l|c|c|l|c|c}
\hline
\multicolumn{2}{c|}{Condition 2} & \multicolumn{3}{c|}{ARI (Agent A)} & \multicolumn{3}{c|}{ARI (Agent B)} & \multicolumn{1}{c|}{$\kappa$} & \multicolumn{1}{c}{ARI (W)}\\ \hline \hline
Pattern & Communication type & Mean & SD & t-test & Mean & SD & t-test & Mean & Mean \\ \hline

\multirow{4}{*}{\shortstack{I: \\Agent A (VSH) \\Agent B (VS)}} & Proposed algorithm & \textbf{0.87} & 0.02 & - & \textbf{\underline{0.84}} & 0.04 & - & \textbf{\underline{0.97}} & 0.93 \\ 
& All acceptance & 0.10 & 0.04 & ** & 0.04 & 0.03 & ** & \textbf{0.17} & - \\
& All rejection & \textbf{\underline{0.88}} & 0.04 & n.s. & \textbf{0.82} & 0.03 & n.s. & 0.00 & - \\ \cline{2-10}
& Integrated model (top line) & 0.86 & 0.02 & n.s. & 0.83 & 0.02 & n.s. & - & - \\ \hline

\multirow{4}{*}{\shortstack{II: \\Agent A (VSH) \\Agent B (SH)}} & Proposed algorithm & \textbf{\underline{0.88}} & 0.03 & - & \textbf{\underline{0.79}} & 0.03 & - & \textbf{\underline{0.97}} & 0.91 \\ 
& All acceptance & 0.22 & 0.09 & ** & 0.18 & 0.05 & ** & \textbf{0.25} & - \\
& All rejection & \textbf{0.88} & 0.03 & n.s. & \textbf{0.78} & 0.02 & n.s. & 0.00 & - \\ \cline{2-10}
& Integrated model (top line) & 0.88 & 0.03 & n.s. & 0.78 & 0.03 & n.s. & - & - \\ \hline

\multirow{4}{*}{\shortstack{III: \\Agent A (VSH) \\Agent B (VH)}} & Proposed algorithm & \textbf{\underline{0.86}} & 0.03 & - & \textbf{\underline{0.80}} & 0.03 & - & \textbf{\underline{0.97}} & 0.90 \\ 
& All acceptance & 0.17 & 0.07 & ** & 0.13 & 0.05 & ** & \textbf{0.23} & - \\
& All rejection & \textbf{0.85} & 0.03 & n.s. & \textbf{0.71} & 0.04 & ** & 0.00 & - \\ \cline{2-10}
& Integrated model (top line) & 0.88 & 0.03 & n.s. & 0.81 & 0.02 & n.s. & - & - \\ \hline
\end{tabular}
}
\label{tab:real_ex_2}
\end{table}

\begin{table}[bp]
\centering
\caption{Results of condition 3 with the ARI means and standard deviations and the means of Kappa coefficients ($\kappa$) in 10 trials with 300 iterations.
In the t-test, **: $(p<0.01)$, *: $(p<0.05)$, n.s.: not significant. The largest and second largest values in the ARIs and Kappa coefficients without the top line are indicated in bold-underlined and bold, respectively. In the patterns, V, S, and H indicate visual, sound, and haptic modalities, respectively.}
\scalebox{0.8}{
\begin{tabular}{c|l|c|c|l|c|c|l|c|c}
\hline
\multicolumn{2}{c|}{Condition 3} & \multicolumn{3}{c|}{ARI (Agent A)} & \multicolumn{3}{c|}{ARI (Agent B)} & \multicolumn{1}{c|}{$\kappa$} & \multicolumn{1}{c}{ARI (W)}\\ \hline \hline
Pattern & Communication type & Mean & SD & t-test & Mean & SD & t-test & Mean & Mean \\ \hline
\multirow{4}{*}{\shortstack{I: \\Agent A (VSH) \\Agent B (V)}} & Proposed algorithm & \textbf{0.87} & 0.03 & - & \textbf{\underline{0.74}} & 0.04 & - & \textbf{\underline{0.98}} & 0.87 \\ 
& All acceptance & 0.04 & 0.02 & ** & 0.01 & 0.01 & ** & \textbf{0.05} & - \\
& All rejection & \textbf{\underline{0.87}} & 0.03 & n.s. & \textbf{0.37} & 0.03 & ** & 0.00 & - \\ \cline{2-10}
& Integrated model (top line) & 0.87 & 0.03 & n.s. & 0.75 & 0.02 & n.s. & - & - \\ \hline
\multirow{4}{*}{\shortstack{II: \\Agent A (VSH) \\Agent B (S)}} & Proposed algorithm & \textbf{0.86} & 0.04 & - & \textbf{\underline{0.72}} & 0.04 & - & \textbf{\underline{0.97}} & 0.91\\ 
& All acceptance & 0.07 & 0.02 & ** & 0.03 & 0.01 & ** & \textbf{0.10} & - \\
& All rejection & \textbf{\underline{0.87}} & 0.03 & n.s. & \textbf{0.44} & 0.05 & ** & -0.00 & - \\ \cline{2-10}
& Integrated model (top line) & 0.88 & 0.04 & n.s. & 0.73 & 0.03 & n.s. & - & - \\ \hline
\multirow{4}{*}{\shortstack{III: \\Agent A (VSH) \\Agent B (H)}} & Proposed algorithm & \textbf{\underline{0.88}} & 0.03 & - & \textbf{\underline{0.49}} & 0.04 & - & \textbf{\underline{0.94}} & 0.75 \\ 
& All acceptance & 0.15 & 0.08 & ** & 0.20 & 0.05 & ** & \textbf{0.26} & - \\
& All rejection & \textbf{0.87} & 0.04 & n.s. & \textbf{0.24} & 0.02 & ** & -0.00 & - \\ \cline{2-10}
& Integrated model (top line) & 0.88 & 0.03 & n.s. & 0.49 & 0.04 & n.s. & - & - \\ \hline
\end{tabular}
}
\label{tab:real_ex_3}
\end{table}

\begin{table}[bp]
\centering
\caption{Results of condition 4 with the ARI means and standard deviations and the means of Kappa coefficients ($\kappa$) in 10 trials with 300 iterations.
In the t-test, **: $(p<0.01)$, *: $(p<0.05)$, n.s.: not significant. The largest and second largest values in the ARIs and Kappa coefficients without the top line are indicated in bold-underlined and bold, respectively. In the patterns, V, S, and H indicate visual, sound, and haptic modalities, respectively.}
\scalebox{0.8}{
\begin{tabular}{c|l|c|c|l|c|c|l|c|c}
\hline
\multicolumn{2}{c|}{Condition 4} & \multicolumn{3}{c|}{ARI (Agent A)} & \multicolumn{3}{c|}{ARI (Agent B)} & \multicolumn{1}{c|}{$\kappa$} & \multicolumn{1}{c}{ARI (W)}\\ \hline \hline
Pattern & Communication type & Mean & SD & t-test & Mean & SD & t-test & Mean & Mean \\ \hline
\multirow{4}{*}{\shortstack{I: \\Agent A (VS) \\Agent B (H)}} & Proposed algorithm & \textbf{\underline{0.85}} & 0.01 & - & \textbf{\underline{0.49}} & 0.04 & - & \textbf{\underline{0.95}} & 0.74 \\ 
& All acceptance & 0.00 & 0.00 & ** & 0.05 & 0.01 & ** & \textbf{0.08} & - \\
& All rejection & \textbf{0.83} & 0.04 & n.s. & \textbf{0.24} & 0.01 & ** & -0.00 & - \\ \cline{2-10}
& Integrated model (top line) & 0.84 & 0.03 & n.s. & 0.50 & 0.03 & n.s. & - & - \\ \hline

\multirow{4}{*}{\shortstack{II: \\Agent A (SH) \\Agent B (V)}} & Proposed algorithm & \textbf{\underline{0.80}} & 0.04 & - & \textbf{\underline{0.71}} & 0.03 & - & \textbf{\underline{0.97}} & 0.88 \\ 
& All acceptance & 0.06 & 0.01 & ** & 0.01 & 0.01 & ** & \textbf{0.03} & - \\
& All rejection & \textbf{0.79} & 0.03 & n.s. & \textbf{0.34} & 0.02 & ** & -0.00 & - \\ \cline{2-10}
& Integrated model (top line) & 0.81 & 0.01 & n.s. & 0.71 & 0.02 & n.s. & - & - \\ \hline

\multirow{4}{*}{\shortstack{III: \\Agent A (VH) \\Agent B (S)}} & Proposed algorithm & \textbf{\underline{0.82}} & 0.04 & - & \textbf{\underline{0.71}} & 0.05 & - & \textbf{\underline{0.96}} & 0.89 \\ 
& All acceptance & 0.04 & 0.02 & ** & 0.01 & 0.01 & ** & \textbf{0.05} & - \\
& All rejection & \textbf{0.73} & 0.04 & ** & \textbf{0.41} & 0.04 & ** & 0.00 & - \\
\cline{2-10} 
& Integrated model (top line) & 0.82 & 0.03 & n.s. & 0.73 & 0.02 & n.s. & - & - \\ \hline
\end{tabular}
}
\label{tab:real_ex_4}
\end{table}

\begin{table}[htbp]
\centering
\caption{ARIs of categories in DM or MDM with the patterns of modalities using real-world datasets for each agent}
\scalebox{0.8}{
\begin{tabular}{c|c|c}
\hline
\multicolumn{1}{c}{DM or MDM} & \multicolumn{1}{|c}{ARI (Agent A)} & \multicolumn{1}{|c}{ARI (Agent B)}\\ \hline \hline
Pattern & Mean & Mean\\ \hline 
Vision & 0.38 & 0.36 \\
Sound & 0.41 & 0.42 \\
Haptic & 0.24 & 0.25 \\
Vision，Sound & 0.83 & 0.83 \\
Sound，Haptic & 0.78 & 0.77 \\
Vision，Haptic & 0.72 & 0.72 \\
Vision，Sound，Haptic & \textbf{0.87} & \textbf{0.86}\\ \hline
\end{tabular}
}
\label{tab:real_ex_sub}
\end{table}

\subsection{Discussion}
In condition 1, we verified hypothesis (1) for Inter-MDM, that the model enables the agents to form perceptual categories and share signs based on multimodal sensory information.
In the results of condition 1 in Table~\ref{tab:real_ex_1}, because the ARIs of agents A and B in the proposed algorithm are not significantly different from those of the top line, the formation of perceptual categories based on multimodal information was confirmed. From the value of the kappa coefficient in the proposed algorithm, the sharing of signs between agents based on multimodal information was also confirmed.

In conditions 2 to 4, we verified hypothesis (2) that the accuracy of categorization in an agent is improved by semiotic communication with another agent even when some modality information is missing.
In Table~\ref{tab:real_ex_2} for condition 2, where one modality of agent B is missing, it can be confirmed that the proposed algorithm obtains a higher ARI (agent B) than all rejection in which categories are formed without communication. 
In Table~\ref{tab:real_ex_3} for condition 3, in which two modalities of agent B are missing, the difference in ARI (agent B) between the proposed algorithm and all rejection is even larger than that in condition 1.
In Table~\ref{tab:real_ex_4} for condition 4, in which modalities of agents A and B are missing, it can be confirmed through comparison with all rejection that the ARIs of both agents were improved in the proposed algorithm.
From the results of conditions 2 to 4, it was clarified that semiotic communication with other agents compensates for the missing modality of an agent and improves the accuracy of categorization even when using real-world datasets.
	
In contrast, as presented in Table~\ref{tab:real_ex_2} for condition 2, no significant difference was obtained between the proposed algorithm and all rejection regarding the ARI of categories of agent B in the cases of patterns 1 and 2. A similar tendency is confirmed in the ARI of agent A in Table~\ref{tab:real_ex_4} for condition 4.
Ancillary experiments were performed to determine why the results differed from those for the synthetic data in patterns 1 and 2.
Table~\ref{tab:real_ex_sub} presents the ARIs of categories in agents A and B, which classify real-world data by DM using one modality data, and by MDM using two and three modality data.
Because the ARI of agent B with vision and sound was 0.83 and the ARI of agent B with sound and haptics was 0.77, it can be observed that agent B performs highly accurate categorization using only two modalities in the case of patterns 1 and 2. In the case of visual and haptic information corresponding to pattern 3, the ARI of agent B was 0.72, and the ARI in the case of vision, sound, and haptics was 0.86. In pattern 3, a significant difference is presumed because this difference is larger than that in patterns 1 and 2. Thus, the differences from the synthetic data results are considered to be due to the characteristics of the modalities of the real-world datasets used in the experiment. For the synthetic datasets, observations were generated by setting the characteristics of the three modalities that differed for each category. Therefore, observations of three modalities were required for highly accurate categorization. However, because real-world datasets composed of observations of real objects do not have settings, as in the synthetic datasets, it is presumed that highly accurate categorization was achieved with only two modalities (e.g., vision and haptics).

\section{Experiment 3: Interpersonal cross-modal inference}\label{sec:6}
\label{sec:crossmodal}
To verify whether Inter-MDM allows cross-modal inference between agents, we performed an experiment in which one agent predicted unobserved information from a sign sampled from another agent.

\subsection{Conditions}
In the experiment, the parameters of the model were inferred under the condition that both agents had three modalities and the interpersonal cross-modal inference was performed based on the learned parameters.
Specifically, a sign ($w_d^A)$ was sampled based on the observed information ($o_d^A$) and parameters (${\bf{\Phi}}^A$) of agent A, and unobserved information ($\hat{o_d^B}$) of agent B was predicted based on the sampled sign ($w_d^A$). The dataset used for training was the real-world dataset used in Experiment 2.

\subsection{Evaluation criteria}
Cosine similarity and Jensen--Shannon divergence (JSD), described in Appendix~\ref{sec:Cos and JSD}, were adopted as the criteria for evaluating the similarity between the predicted information ($\hat{o_{v,d}^{B}},\hat{o_{s,d}^{B}},\hat{o_{h,d}^{B}}$) and actual observations.

The cosine similarity is a criterion that represents the closeness of the angles formed by vectors in a vector space. The value of cosine similarity increases when the vectors are similar, and the maximum value is 1.0. The histogram of the observed information is regarded as a vector, and the cosine similarity is calculated.
The average value of cosine similarity was calculated using the following formula:
\begin{eqnarray}
\label{eq:cos_sim_AB}
\bar{ cos}({\bf{O}}_{*}^{A},\hat{{\bf{O}}_{*}^{B}}) = \frac{\sum_{d} cos (o_{*,d}^{A},\hat{o_{*,d}^{B}})}{D},
\end{eqnarray}
where $D$ is the number of data points, $o_{*,d}^{A}$ is an observation of agent A, $\hat{o_{*,d}^{B}}$ is a predicted observation of agent B, and ${\bf{O}}_{*}^{A}$ and $\hat{{\bf{O}}_{*}^{B}}$ are sets of $o_{*,d}^{A}$ and $\hat{o_{*,d}^{B}}$, respectively. 

JSD is a criterion that measures the similarity between probability distributions via the Kullback–-Leibler divergence (KLD). The value decreases when the probability distributions are similar, and the minimum value is 0.0.
The average value of JSD was calculated using the following formula:
\begin{eqnarray}
\label{eq:jsd_AB}
\bar{D_{JS}}({{\bf{O}}_{*}^{A}}^{\prime},\hat{{\bf{O}}_{*}^{B}}^{\prime}) = \frac{\sum_{d} D_{JS}({o_{*,d}^{A}}^{\prime},\hat{o_{*,d}^{B}}^{\prime})}{D},
\end{eqnarray}
where $D$ is the number of data points, ${o_{*,d}^{A}}^{\prime}$ is the probability distribution obtained by normalizing the histogram of the observed information ($o_{*,d}^{A}$) of agent A, $\hat{o_{*,d}^{B}}^{\prime}$ is the probability distribution obtained by normalizing the histogram of the observed information ($\hat{o_{*,d}^{B}}$) predicted by agent B, and ${{\bf{O}}_{*}^{A}}^{\prime}$ and $\hat{{\bf{O}}_{*}^{B}}^{\prime}$ are sets of ${o_{*,d}^{A}}^{\prime}$ and $\hat{o_{*,d}^{B}}^{\prime}$, respectively.

\subsection{Experimental results}
\begin{figure}[ht]
\begin{center}			
\includegraphics[scale=0.8]{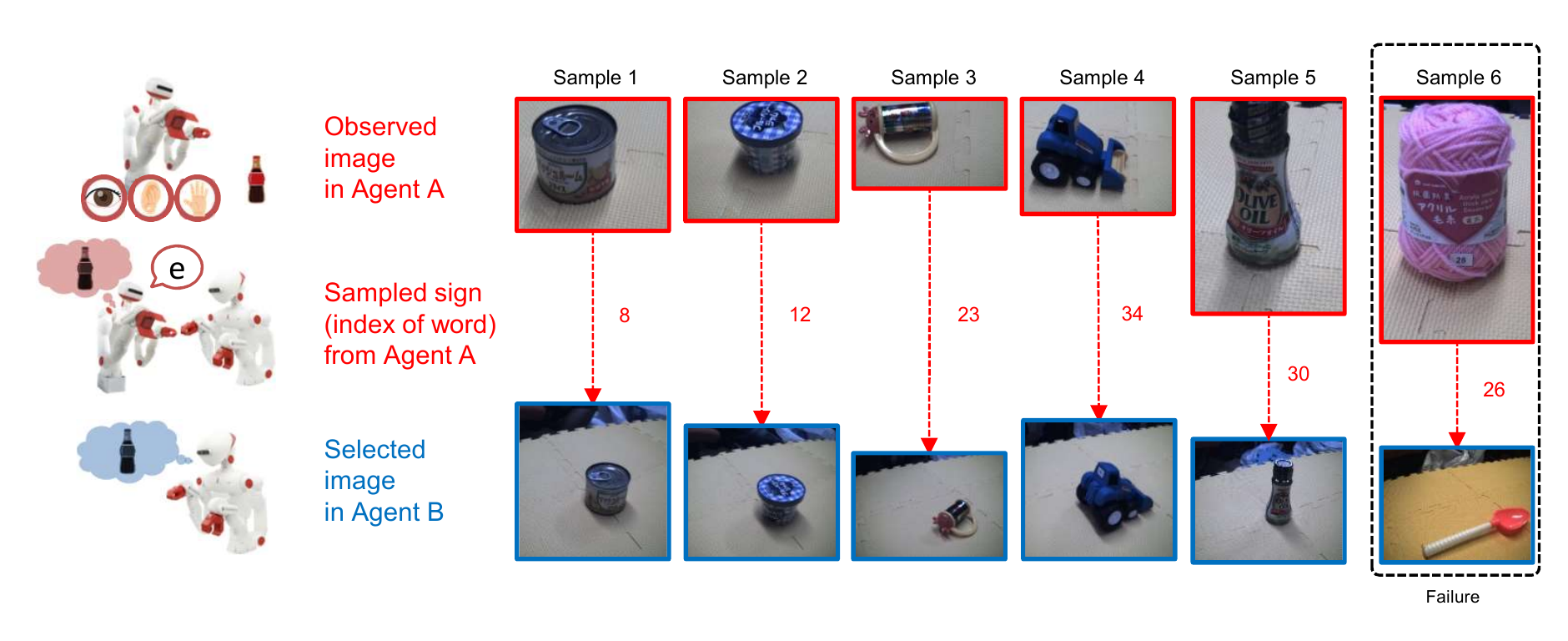}
\caption{Samples of the image prediction in agent B from a sign suggested by agent A. An image in agent B is selected based on the JSD from the predicted observation $\hat{o_{v,d}^{B}}$ in agent B}
\label{fig:JSD_sample}
\end{center}
\end{figure}

As a qualitative evaluation, an experiment on interpersonal cross-modal inference was performed using real-world datasets with the following process:
\begin{enumerate}
 \item Agent A obtains the observation ($o_d^A$) of an object.
 \item A sign ($w_d$) is sampled from the observation ($o_d^A$) and learned parameters by Inter-MDM in agent A.
 \item Agent B predicts a visual observation ($\hat{o_{v,d}^{B}}$) from the sign ($w_d$) and the learned parameters by Inter-MDM. 
\end{enumerate}
The experimental results of the qualitative evaluation are presented in Figure~\ref{fig:JSD_sample}.
The image surrounded by the red line is the image observed by agent A, and the number next to the arrow is a sign (the index of a word) sampled from agent A. The image surrounded by the blue line is the image selected by agent B from the dataset based on the sampled sign.
This image is selected from the observed images in the dataset of agent B with the distribution of features closest to the visual observation ($\hat{o_{v,d}^{B}}$) predicted by agent B. JSD was adopted as the criterion for calculating the closeness of the distribution of image features.
In the samples of six objects, samples 1--5 are success cases in which agent B can predict an image of the same object as the object observed by agent A, and sample 6 is a failure case in which a different object is predicted. This demonstrates that agent B can predict and select the image features of the same object based on the signs sampled from agent A in Inter-MDM. Failure cases can be considered as a few cases triggered by the probabilistic generation of categories and signs.

Next, a quantitative evaluation of the similarities and differences in observations between agents is described. In the quantitative evaluation, the similarity and difference between the observations of agent A and the observations predicted by agent B were evaluated in each modality in terms of cosine similarity and JSD, respectively. In addition, the models were trained and evaluated using three communication types (i.e., the proposed algorithm, all acceptance, and all rejection) for comparison.

Table~\ref{tab:cos_JSD_result} presents the results of quantitative evaluation using cosine similarity and JSD. The average values of the cosine similarity and JSD for 400 data points with three modalities (i.e., vision, sound, and haptics) in the three communication types are listed.
The experimental results of cosine similarity in Table~\ref{tab:cos_JSD_result} demonstrate that the proposed algorithm enables agent B to predict the unobserved information that is most similar to the observed information of agent A.

The experimental results of JSD in Table~\ref{tab:cos_JSD_result} demonstrate that the proposed algorithm enables agent B to predict the unobserved information that has the least difference from the observed information of agent A.
The results of the quantitative evaluation demonstrate that the communication via the proposed algorithm enables more accurate cross-modal inference between agents than that in other algorithms (i.e., all acceptance and all rejection).

\begin{table}[htbp]
\centering
\caption{Cosine similarity and JSD between the observation ($o_{*,d}^{A}$) of agent A and the predicted observation ($\hat{o_{*,d}^{B}}$) in agent B. $*$ is a modality (i.e., $v,s,h$). Cos and JSD indicate the means of the cosine similarity and JSD values, respectively. The largest values in cosine similarity and the smallest values in JSD are indicated in bold and underlined, respectively.}
\scalebox{0.9}{
\begin{tabular}{c|l|c|c|c|c|c|c}
\hline
\multicolumn{2}{c|}{modalities} & \multicolumn{2}{c|}{Vision} & \multicolumn{2}{c|}{Sound} & \multicolumn{2}{c}{Haptic} \\ \hline \hline
Pattern & Communication type & Cos ($\uparrow$) & JSD ($\downarrow$) & Cos($\uparrow$) & JSD ($\downarrow$) & Cos ($\uparrow$) & JSD ($\downarrow$) \\ \hline
 \multirow{3}{*}{\shortstack{Agent A (VSH), \\Agent B (VSH)}} & Proposed algorithm & \textbf{\underline{0.31}} & \textbf{\underline{0.42}} & \textbf{\underline{0.63}} & \textbf{\underline{0.19}} & \textbf{\underline{0.89}} & \textbf{\underline{0.06}}  \\ 
 & All acceptance & 0.24 & 0.47 & 0.52 & 0.24 & 0.73 & 0.21 \\
 & All rejection & 0.16 & 0.54 & 0.38 & 0.36 & 0.31 & 0.45 \\ \hline
\end{tabular}
}
\label{tab:cos_JSD_result}
\end{table}

\end{CJK}

\section{Conclusions}\label{sec:7}
\begin{CJK}{UTF8}{min}

In this study, we proposed Inter-MDM, which models the symbol emergence system where each agent has multimodal sensor systems, forms multimodal object categories, and shares signs with another agent.
Inter-MDM was modeled as a PGM, a multimodal extension of Inter-DM~\cite{Hagiwara19}, which models a symbol emergence system comprising two agents with a visual modality.

From experiments using a multimodal dataset acquired by a robot in the real world, we clarified whether Inter-MDM can realize the functions proposed in the following questions.
\begin{enumerate}
 \item Is it possible to extend Inter-DM to make it multimodal and realize symbol emergence based on categorizing multimodal sensory information?
 \item Is it possible to improve the categorization accuracy in an agent via semiotic communication with another agent, even if some modalities are missing?
 \item Is it possible to infer unobserved modality information based on a sign shared between two agents in a bottom-up manner?
\end{enumerate}

In Experiment 1, using a synthetic dataset, and Experiment 2, using a real-world dataset, questions (1) and (2) were verified based on the categorization accuracy and coincidence of signs between agents by comparing with baseline algorithms (i.e., all acceptance and all rejection) and a top line algorithm (i.e., integrated model).
From the experimental results for condition 1, we clarified that Inter-MDM enables two agents to form perceptual categories and share signs based on multimodal sensory information.
From the experimental results for conditions 2 to 4 with different sets of modalities, we clarified that Inter-MDM can improve the accuracy of categorization in an agent by semiotic communication with another agent, even if some modalities are missing.
In Experiment 3 for interpersonal cross-modal inference, we clarified that Inter-MDM can enable an agent to infer unobserved sensory modality information (i.e., visual, sound, and haptic) based on a sign shared between two agents.

In this study, we clarified that Inter-MDM realizes the functions in the above-mentioned questions from the experimental results and developed a computational model that represents the symbol emergence system between two agents, where each agent has multimodal sensor systems, forms multimodal object categories, and shares signs with the other agent.

In future work, we plan to perform experiments on naming games between humans and robots and elucidate the mechanism of symbol emergence between humans.
To prepare for these experiments, we are performing the following challenging tasks:
\begin{enumerate}
    \item Extension of Inter-MDM from the exchange of a sign to the exchange of a sentence. This extension is implemented by applying a simultaneous learning model of words and concepts~\cite{Araki12} and the multi-channel categorization model ~\cite{Taniguchi17} to Inter-MDM.
    \item Extension of Inter-MDM to a model that can generate images directly by applying the deep generative model~\cite{Kingma14}, the neural topic model~\cite{Srivastava17}, and Neuro-SERKET~\cite{Taniguchi20}.
    \item Extension of Inter-MDM from two agents to a naming game involving three or more agents. The interpersonal inference based on the M-H algorithm cannot be directly applied to the case of three or more agents. Communication protocols and mathematical interpretations should be considered in this case.
\end{enumerate}

\end{CJK}

\section*{Acknowledgments}
This study was partially supported by the Japan Science and Technology Agency (JST) Core Research for Evolutionary Science and Technology (CREST) research program, under Grant JPMJCR15E3, by the Japan Society for the Promotion of Science (JSPS) KAKENHI under Grant JP18K18134, and by MEXT Grant-in-Aid for Scientific Research on Innovative Areas 4903 (Co-creative Language Evolution), 17H06383.

\label{lastpage}

\bibliographystyle{tADR}
\bibliography{tADR}

\clearpage

%\begin{comment}

\appendices
\section{Derivation process of an acceptance rate $z$ in Algorithm~\ref{alg:alg3}}
This section describes the derivation process of the acceptance rate ($z$) on the fifteenth line of Algorithm~\ref{alg:alg3}. 
The formulas for the acceptance rate in the M-H algorithm are defined as follows:
\begin{eqnarray}
Z(s,s^*)&=&{\rm{min}}(1,z) \\
z&=&\frac{P(s^*)Q(s|s^*)}{P(s)Q(s^*|s)},
\end{eqnarray}
where $s$ and $s^*$ are the past and new samples, respectively. $P(\cdot)$ and $Q(\cdot)$ are the target and proposed distributions, respectively.
The purpose of the M-H algorithm is to generate a sample according to the target distribution.

According to the above formulas, the acceptance rate $z$ in Algorithm~\ref{alg:alg2} is derived. In Algorithm~\ref{alg:alg2}, if the past and new samples are $w_d^{Li}$ and $w_d^{Sp}$, respectively, and the target distribution and proposed distribution are $P(w_d|c_d^{Sp},c_d^{Li},{\bf{\Theta}}^{Sp},{\bf{\Theta}}^{Li})$ and $P(w_d|c_d^{Sp},{\bf{\Theta}}^{Sp})$, respectively, the following equation is obtained:
\begin{eqnarray}
\label{eq:mh_z_1}
z=\frac{P(w_d^{Sp}|c_d^{Sp},c_d^{Li},{\bf{\Theta}}^{Sp},{\bf{\Theta}}^{Li})
P(w_d^{Li}|c_d^{Sp},{\bf{\Theta}}^{Sp})}
{P(w_d^{Li}|c_d^{Sp},c_d^{Li},{\bf{\Theta}}^{Sp},{\bf{\Theta}}^{Li})
P(w_d^{Sp}|c_d^{Sp},{\bf{\Theta}}^{Sp})} 
\end{eqnarray}
The first term of the numerator of Equation(\ref{eq:mh_z_1}) using Bayes' theorem is as follows:
\begin{eqnarray}
P(w_d^{Sp}|c_d^{Sp},c_d^{Li},{\bf{\Theta}}^{Sp},{\bf{\Theta}}^{Li})
&=&\frac{P(c_d^{Sp}|c_d^{Li},{\bf{\Theta}}^{Sp},{\bf{\Theta}}^{Li},w_d^{Sp})
P(w_d^{Sp}|c_d^{Li},{\bf{\Theta}}^{Sp},{\bf{\Theta}}^{Li})}
{P(c_d^{Sp}|c_d^{Li},{\bf{\Theta}}^{Sp},{\bf{\Theta}}^{Li})} \nonumber \\
&\propto&P(c_d^{Sp}|c_d^{Li},{\bf{\Theta}}^{Sp},{\bf{\Theta}}^{Li},w_d^{Sp})
P(w_d^{Sp}|c_d^{Li},{\bf{\Theta}}^{Sp},{\bf{\Theta}}^{Li}) \nonumber \\
\label{eq:mh_z_2}
\end{eqnarray}
The first term of Equation (\ref{eq:mh_z_2}) is transformed using Bayes' theorem and the Markov blanket as follows:
\begin{eqnarray}
P(c_d^{Sp}|c_d^{Li},{\bf{\Theta}}^{Sp},{\bf{\Theta}}^{Li},w_d^{Sp})
&=&\frac{P({\bf{\Theta}}^{Li}|c_d^{Sp},c_d^{Li},{\bf{\Theta}}^{Sp},w_d^{Sp})
P(c_d^{Sp}|c_d^{Li},{\bf{\Theta}}^{Sp},w_d^{Li})}
{P({\bf{\Theta}}^{Li}|c_d^{Li},{\bf{\Theta}}^{Sp},w_d^{Sp})} \nonumber \\
&\propto&\frac{P({\bf{\Theta}}^{Li}|c_d^{Li},w_d^{Sp})
P(c_d^{Sp}|{\bf{\Theta}}^{Sp},w_d^{Sp})}
{P({\bf{\Theta}}^{Li}|c_d^{Li},w_d^{Sp})} \nonumber \\
&=&P(c_d^{Sp}|{\bf{\Theta}}^{Sp},w_d^{Sp})
\label{eq:mh_z_3}
\end{eqnarray}
The latter term of Equation (\ref{eq:mh_z_2}) is transformed using Bayes' theorem and the Markov blanket as follows:
\begin{eqnarray}
P(w_d^{Sp}|c_d^{Li},{\bf{\Theta}}^{Sp},{\bf{\Theta}}^{Li})
&=&\frac{P(c_d^{Li}|{\bf{\Theta}}^{Sp},{\bf{\Theta}}^{Li},w_d^{Sp})
P(w_d^{Sp}|{\bf{\Theta}}^{Sp},{\bf{\Theta}}^{Li})}
{P(c_d^{Li}|{\bf{\Theta}}^{Sp},{\bf{\Theta}}^{Li})} \nonumber \\
&\propto&\frac{P(c_d^{Li}|{\rm \bf{\Theta}}^{Li},w_d^{Sp})P(w_d^{Sp})}
{P(c_d^{Li}|{\bf{\Theta}}^{Li})} \nonumber \\
&\propto&P(c_d^{Li}|{\bf{\Theta}}^{Li},w_d^{Sp})P(w_d^{Sp})
\label{eq:mh_z_4}
\end{eqnarray}
Substituting Equations (\ref{eq:mh_z_3}) and (\ref{eq:mh_z_4}) into Equation (\ref{eq:mh_z_2}) yields the following equation:
\begin{eqnarray}
P(w_d^{Sp}|c_d^{Sp},c_d^{Li},{\bf{\Theta}}^{Sp},{\bf{\Theta}}^{Li})
\propto P(c_d^{Sp}|{\bf{\Theta}}^{Sp},w_d^{Sp})
P(c_d^{Li}|{\bf{\Theta}}^{Li},w_d^{Sp})P(w_d^{Sp})
\label{eq:mh_z_5}
\end{eqnarray}
The latter term of the numerator of Equation (\ref{eq:mh_z_1}) is transformed using Bayes' theorem as follows:
\begin{eqnarray}
P(w_d^{Li}|c_d^{Sp},{\bf{\Theta}}^{Sp})
&=&\frac{P(c_d^{Sp}|{\bf{\Theta}}^{Sp},w_d^{Li})
P(w_d^{Li}|{\bf{\Theta}}^{Sp})}{P(c_d^{Sp}|{\bf{\Theta}}^{Sp})} \nonumber \\
&\propto&P(c_d^{Sp}|{\bf{\Theta}}^{Sp},w_d^{Li})P(w_d^{Li})
\label{eq:mh_z_6}
\end{eqnarray}
The following equation for the acceptance rate $z$ is obtained by transforming Equation (\ref{eq:mh_z_1}) using Equations (\ref{eq:mh_z_5}) and (\ref{eq:mh_z_6}).
\begin{eqnarray}
\label{eq:mh_z_7}
z&=&\frac{P(w_d^{Sp}|c_d^{Sp},c_d^{Li},{\bf{\Theta}}^{Sp},{\bf{\Theta}}^{Li})
P(w_d^{Li}|c_d^{Sp},{\bf{\Theta}}^{Sp})}
{P(w_d^{Li}|c_d^{Sp},c_d^{Li},{\bf{\Theta}}^{Sp},{\bf{\Theta}}^{Li})
P(w_d^{Sp}|c_d^{Sp},{\bf{\Theta}}^{Sp})} \nonumber \\
&\propto&\frac{P(c_d^{Sp}|{\bf{\Theta}}^{Sp},w_d^{Sp})
P(c_d^{Li}|{\bf{\Theta}}^{Li},w_d^{Sp})P(w_d^{Sp})
P(c_d^{Sp}|{\bf{\Theta}}^{Sp},w_d^{Li})P(w_d^{Li})}{
P(c_d^{Sp}|{\bf{\Theta}}^{Sp},w_d^{Li})
P(c_d^{Li}|{\bf{\Theta}}^{Li},w_d^{Li})P(w_d^{Li})
P(c_d^{Sp}|{\bf{\Theta}}^{Sp},w_d^{Sp})P(w_d^{Sp})} \nonumber \\
&=&\frac{P(c_d^{Li}|{\bf{\Theta}}^{Li},w_d^{Sp})}
{P(c_d^{Li}|{\bf{\Theta}}^{Li},w_d^{Li})}
\end{eqnarray}

\section{Gibbs sampling for Inter-MDM}
\label{sec:Gibbs sampling}
Gibbs sampling is an M-H algorithm that samples according to a probability distribution obtained from a conditional probability distribution marginalized from a joint distribution. This algorithm has a guaranteed solution as an approximate inference based on MCMC. 
Algorithm~\ref{alg:Gibbs sampling} presents the Gibbs sampling inference algorithm for the multiagent multimodal symbol emergence model illustrated in Fig.~\ref{fig:model}). 

$i$ denotes the number of iterations. $\rm{\bf{O}}_{*}^A$ and $\rm{\bf{O}}_{*}^B$ are the observations of agents A and B, respectively. $\rm{\bf{C}}^A$ and $\rm{\bf{C}}^B$ are the set of categories in agents A and B, respectively. $\rm{\bf{W}}$ denotes the set of signs.

\begin{algorithm}
\caption{Gibbs sampling algorithm}
\label{alg:Gibbs sampling}
\begin{algorithmic}[1]
\STATE Initialize all parameters 
\FOR {$i=1$ to $I$}
\FOR {$l=1$ to $L$}
\STATE  $\phi_{v,l}^{A[i]} \sim {\rm Dir}(\phi_{v,l}^{A[i]}|{\rm \bf{O}}_v^A,{\rm \bf{C}}^{A[i-1]},\beta_{v}) $
\STATE  $\phi_{s,l}^{A[i]} \sim {\rm Dir}(\phi_{s,l}^{A[i]}|{\rm \bf{O}}_s^A,{\rm \bf{C}}^{A[i-1]},\beta_{s}) $
\STATE  $\phi_{h,l}^{A[i]} \sim {\rm Dir}(\phi_{h,l}^{A[i]}|{\rm \bf{O}}_h^A,{\rm \bf{C}}^{A[i-1]},\beta_{h}) $  
\STATE  $\phi_{v,l}^{B[i]} \sim {\rm Dir}(\phi_{v,l}^{B[i]}|{\rm \bf{O}}_v^B,{\rm \bf{C}}^{B[i-1]},\beta_{v}) $
\STATE  $\phi_{s,l}^{B[i]} \sim {\rm Dir}(\phi_{s,l}^{B[i]}|{\rm \bf{O}}_s^B,{\rm \bf{C}}^{B[i-1]},\beta_{s}) $
\STATE  $\phi_{h,l}^{B[i]} \sim {\rm Dir}(\phi_{h,l}^{B[i]}|{\rm \bf{O}}_h^B,{\rm \bf{C}}^{B[i-1]},\beta_{h}) $  
\ENDFOR
\FOR {$k=1$ to $K$}
\STATE  $\theta_{k}^{A[i]} \sim {\rm Dir}(\theta_{k}^{A[i]}|{\rm \bf{C}}^{A[i-1]},{\rm \bf{W}}^{[i-1]},\alpha) $
\STATE  $\theta_{k}^{B[i]} \sim {\rm Dir}(\theta_{k}^{B[i]}|{\rm \bf{C}}^{B[i-1]},{\rm \bf{W}}^{[i-1]},\alpha) $
\ENDFOR
\FOR {$d=1$ to $D$}
\STATE  $c_{d}^{A[i]} \sim 
{\rm Cat}(c_{d}^{A[i]}|\theta_{w_d^{[i-1]}}^{A[i]})
{\rm Multi}(o_{v,d}^A|\phi_{v,c_d^{A[i]}}^{A[i]})
{\rm Multi}(o_{s,d}^A|\phi_{s,c_d^{A[i]}}^{A[i]})    
{\rm Multi}(o_{h,d}^A|\phi_{h,c_d^{A[i]}}^{A[i]}) $
\STATE  $c_{d}^{B[i]} \sim 
{\rm Cat}(c_{d}^{B[i]}|\theta_{w_d^{[i-1]}}^{B[i]})
{\rm Multi}(o_{v,d}^B|\phi_{v,c_d^{B[i]}}^{B[i]})
{\rm Multi}(o_{s,d}^B|\phi_{s,c_d^{B[i]}}^{B[i]})    
{\rm Multi}(o_{h,d}^B|\phi_{h,c_d^{B[i]}}^{B[i]}) $
\STATE  $w_{d}^{[i]} \sim
{\rm Cat}(c_{d}^{A[i]}|\theta_{w_d^{[i]}}^{A[i]})
{\rm Cat}(c_{d}^{B[i]}|\theta_{w_d^{[i]}}^{B[i]}) $
\ENDFOR
\ENDFOR
\end{algorithmic} 
\end{algorithm}

The derivation process of each line of Gibbs sampling in Algorithm~\ref{alg:Gibbs sampling} is as follows:

In the parameters $\phi_{*,l}^A$ and $\phi_{*,l}^B$ of multinomial distributions,
\begin{eqnarray}
\phi_{*,l}^{A} &\sim& P(\phi_{*,l}^{A}|{\rm \bf{O}}_*^{A},{\rm \bf{C}}^{A},\beta_{*}) \nonumber \\
&\propto& \prod_{d=1}^{D}{\rm Multi}(o_{d}^{A}|\phi_{*,l}^{A}){\rm Dir}(\phi_{*,l}^{A}|\beta_{*}) \nonumber \\
&\propto& {\rm Dir}(\phi_{*,l}^{A}|{\rm \bf{O}}_*^{A},{\rm \bf{C}}^{A},\beta_{*}), \\ \nonumber \\
\phi_{*,l}^{B} &\sim& P(\phi_{*,l}^{B}|{\rm \bf{O}}_*^{B},{\rm \bf{C}}^{B},\beta_{*}) \nonumber \\
&\propto& \prod_{d=1}^{D}{\rm Multi}(o_{d}^{B}|\phi_{*,l}^{B}){\rm Dir}(\phi_{*,l}^{B}|\beta_{*}) \nonumber \\
&\propto& {\rm Dir}(\phi_{*,l}^{B}|{\rm \bf{O}}_*^{B},{\rm \bf{C}}^{B},\beta_{*}),
\end{eqnarray}
where $*$ is $v$, $s$, and $h$, respectively.

In the parameters $\theta_{k}^{A}$ and $\theta_{k}^{B}$ of categorical distributions,
\begin{eqnarray}
\theta_{k}^{A} &\sim& P(\theta_{k}^{A}|{\rm \bf{C}}^{A},{\rm \bf{W}},\alpha) \nonumber \\
&\propto& \prod_{d=1}^{D}{\rm Cat}(c_{d}^{A}|\theta_{k}^{A}){\rm Dir}(\theta_{k}^{A}|\alpha) \nonumber \\
&\propto& {\rm Dir}(\theta_{k}^{A}|{\rm \bf{C}}^{A},{\rm \bf{W}},\alpha), \\ \nonumber \\
\theta_{k}^{B} &\sim& P(\theta_{k}^{B}|{\rm \bf{C}}^{B},{\rm \bf{W}},\alpha) \nonumber \\
&\propto& \prod_{d=1}^{D}{\rm Cat}(c_{d}^{B}|\theta_{k}^{B}){\rm Dir}(\theta_{k}^{B}|\alpha) \nonumber \\
&\propto& {\rm Dir}(\theta_{k}^{B}|{\rm \bf{C}}^{B},{\rm \bf{W}},\alpha).
\end{eqnarray}
In the variables of categories $c_{d}^{A}$, $c_{d}^{B}$, and the sign $w_d$,
\begin{eqnarray}
c_{d}^{A} &\sim& P(c_{d}^{A}|o_{v,d}^{A},o_{s,d}^{A},o_{h,d}^{A},{\rm \bf{\Phi}}_{v}^{A},{\rm \bf{\Phi}}_{s}^{A},{\rm \bf{\Phi}}_{h}^{A},w_{d},{\rm \bf{\Theta}}^{A}) \nonumber \\
&\propto& {\rm Cat}(c_{d}^{A}|\theta_{w_d}^{A}){\rm Multi}(o_{v,d}^{A}|\phi_{v,c_d^A}^{A}){\rm Multi}(o_{s,d}^{A}|\phi_{s,c_d^A}^{A}){\rm Multi}(o_{h,d}^{A}|\phi_{h,c_d^A}^{A}),
\\ \nonumber \\
c_{d}^{B} &\sim& P(c_{d}^{B}|o_{v,d}^{B},o_{s,d}^{B},o_{h,d}^{B},{\rm \bf{\Phi}}_{v}^{B},{\rm \bf{\Phi}}_{s}^{B},{\rm \bf{\Phi}}_{h}^{B},w_{d},{\rm \bf{\Theta}}^{B}) \nonumber \\
&\propto& {\rm Cat}(c_{d}^{B}|\theta_{w_d}^{B}){\rm Multi}(o_{v,d}^{B}|\phi_{v,c_d^B}^{B}){\rm Multi}(o_{s,d}^{B}|\phi_{s,c_d^B}^{B}){\rm Multi}(o_{h,d}^{B}|\phi_{h,c_d^B}^{B}),
\\ \nonumber \\
w_{d} &\sim& P(w_{d}|c_{d}^{A},c_{d}^{B},{\rm \bf{\Theta}}^{A},{\rm \bf{\Theta}}^{B}) \nonumber \\
&\propto& {\rm Cat}(c_{d}^{A}|\theta_{w_d}^{A}){\rm Cat}(c_{d}^{B}|\theta_{w_d}^{B}).
\end{eqnarray}

\section{Cosine similarity and JSD}
\label{sec:Cos and JSD}

The cosine similarity between the $n$-dimensional vectors $u$ ($u=(u_1,...,u_n)$) and $v$ ($v=(v_1,...,v_n)$) is calculated using Equation(\ref{eq:cos_sim}).
\begin{eqnarray}
\label{eq:cos_sim}
{\rm {cos}}(u,v) = \frac{u \cdot v}{|u||v|} = \frac{u_1v_1+ \cdot \cdot \cdot +u_nv_n}{\sqrt{u_1^2+ \cdot \cdot \cdot + u_n^2}\sqrt{v_1^2+ \cdot \cdot \cdot + v_n^2}}
\end{eqnarray}

JSD is calculated by normalizing the fact that the sum of the histogram values of the observed information is 1.0. When $P,Q$ are discrete probability distributions, the KLD for $P$ to $Q$ is calculated using Equation(\ref{eq:kld}).
\begin{eqnarray}
\label{eq:kld}
D_{KL}(P||Q) = \sum_{x} P(x) \log \frac{P(x)}{Q(x)}
\end{eqnarray}

$P(x) and Q(x)$ indicate the probabilities when the values selected according to the probability distributions $P and Q$ are $x$, respectively.
The JSDs of $P$ and $Q$ are calculated using Equation(\ref{eq:jsd}).
\begin{eqnarray}
\label{eq:jsd}
D_{JS}(P||Q) = \frac{1}{2} \{ D_{KL}(P||M) + D_{KL}(Q||M) \} \; \; \; \; (M=\frac{P+Q}{2}),
\end{eqnarray}
where $M$ is the average probability distribution of $P,Q$.

%\end{comment}

\end{document}